%% file: root.tex
\documentclass[letterpaper, 10 pt, conference]{ieeeconf}  

\IEEEoverridecommandlockouts                              

\usepackage{graphics} 
\usepackage{amsmath} 
\usepackage{amssymb}  
\usepackage{amstext}
\usepackage{bm}
\usepackage{mathtools}
\usepackage{cite}
\usepackage{xcolor}
\usepackage{adjustbox}
\usepackage{url}
\usepackage[font=small,tableposition=top]{caption}

\usepackage{glossaries}

\title{\LARGE \bf
Regularized Deep Signed Distance Fields \\ for Reactive Motion Generation
}

\author{Puze Liu, Kuo Zhang, Davide Tateo, Snehal Jauhri, Jan Peters and Georgia Chalvatzaki
\thanks{Computer Science Department, Technische Universit\"at Darmstadt, Germany %
{\tt\footnotesize puze@robot-learning.de, kuo.zhang@stud.tu-darmstadt.de, \{davide.tateo, snehal.jauhri,
jan.peters, georgia.chalvatzaki\} @tu-darmstadt.de}
}
\thanks{This research received funding by the DFG Emmy Noether Programme (\#448644653), the RoboTrust project of the Centre Responsible Digitality Hessen, Germany and the China Scholarship Council (No. 201908080039).
Research presented in this paper has been supported by the German Federal Ministry of Education and Research (BMBF) within a subproject ``Modeling and exploration of the operational area, design of the AI assistance as well as legal aspects of the use of technology'' of the collaborative KIARA project (grant no. 13N16274).}
}

\input{math}
\input{acronyms}

\begin{document}

\newlength{\imagewidth}
\setlength{\imagewidth}{.21\textwidth}

\maketitle
\thispagestyle{empty}
\pagestyle{empty}

\begin{abstract}
Autonomous robots should operate in real-world dynamic environments and collaborate with humans in tight spaces. A key component for allowing robots to leave structured lab and manufacturing settings is their ability to evaluate online and real-time collisions with the world around them. Distance-based constraints are fundamental for enabling robots to plan their actions and act safely, protecting both humans and their hardware. However, different applications require different distance resolutions, leading to various heuristic approaches for measuring distance fields w.r.t. obstacles, which are computationally expensive and hinder their application in dynamic obstacle avoidance use-cases. We propose \gls{redsdf}, a single neural implicit function that can compute smooth distance fields at any scale, with fine-grained resolution over high-dimensional manifolds and articulated bodies like humans, thanks to our effective data generation and a simple \textit{inductive bias} during training. We demonstrate the effectiveness of our approach in representative simulated tasks for \gls{wbc} and safe \gls{hri} in shared workspaces. Finally, we provide proof of concept of a real-world application in a \gls{hri} handover task with a mobile manipulator robot.  
\end{abstract}

\section{Introduction}
Safety is an essential prerequisite for the real-world deployment of robots and their interaction with the dynamic world and the humans in it. The study of safety constraints in robotics that can filter dangerous actions is as long as the field itself. A common way of approaching the problem is to define constraint functions that apply to specific applications and tasks.
The imposed constraints can ensure safety by excluding unsafe parts of the state and action space to avoid collisions and self-collision, harming humans in shared workspaces, or for specifying task-specific workspaces, e.g., in human-robot collaboration, robot-assisted feeding, dressing, etc. Each of these tasks may demand a different set of constraints to be specified and satisfied during deployment.  

\begin{figure}
    \centering
    \includegraphics[width=0.8\columnwidth]{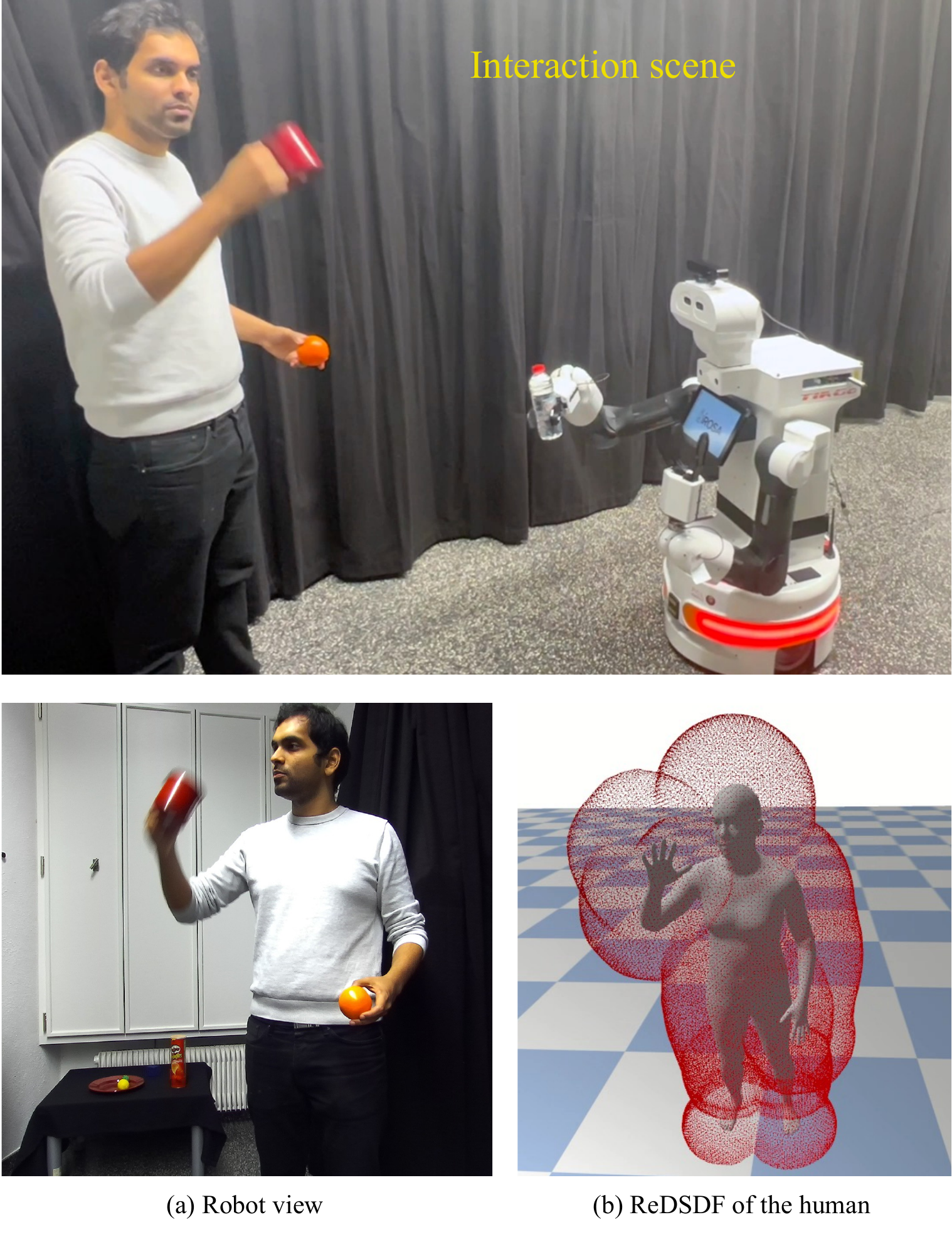}
    \vspace{-0.3em}
    \caption{Real-world evaluation of our learned \gls{redsdf} model in a \gls{hri} scenario. The mobile manipulator robot TIAGo++ has to approach the human and deliver him a water bottle without intervening in the safety boundary. (a) The perceived instance in the robot's field of view. (b) The perceived human manifold, along with the 20cm level set (red point-cloud), i.e. the set of points at 20cm distance to the human.}
    \label{fig:real_world_interaction}
    \vspace{-0.65cm}
\end{figure}

Despite their importance, constraint functions may be difficult to hand-design, while ill-specified constraints may often make it challenging for robots to plan \cite{da2019collision}, optimize \cite{ratliff2009chomp}, learn \cite{liu22robot}, or act reactively \cite{ratliff2018riemannian}. One way of specifying a constraint is to define its boundaries as a level curve $k$ of an implicit function $f(X)=k$, where $X$ can be any query point in space. This formulation is particularly convenient because it defines the constraint over a surface intuitively. Second, it is easy to reason about constraint violations/slack, allowing for easier optimization. Furthermore, if the constraint is also differentiable, it is also possible to use gradient-based projection algorithms to enforce them.

It is common in robotics to formulate constraints in terms of distance w.r.t. a given object. Measuring distances w.r.t. to manifolds is necessary for most robot applications for collision avoidance \cite{barbosa2021risk}, but also for \gls{hri}, where the robot should position itself at a certain distance from the surrounding humans to interact with them safely while respecting social norms \cite{lam2010human,singamaneni2021human, ferraguti2020safety}. 
\gls{sdf} is a prominent representation for expressing distance w.r.t. a given surface, by defining a function that precomputes the distance of an arbitrary point in the space w.r.t. a fixed surface. Every point on the surface has zero distance, while points outside it have a signed distance. The sign is chosen by convention to be positive in the line of sight w.r.t. the sensor detecting the surface. If a surface represents a closed region of space, e.g., an object or a human, we can consider every point inside this region as a negative distance. Different variants of \gls{sdf} were introduced in particular for robotics to counteract the high computational complexity and the non-smooth representations that arise from querying multiple points from numerous static or even dynamic obstacles \cite{finean2021simultaneous}.

In this paper, we present \gls{redsdf}, an approach that extends the concept of \gls{sdf}s for arbitrary articulated objects such as robotic manipulators and humans (Fig. \ref{fig:real_world_interaction}). \gls{redsdf} allows us to define complex distance functions to be used as constraints or energy functions to avoid collisions and achieve safe interactions. 
Our method provides a single deep model that not only preserves structural information in the proximity of articulated bodies, but also effectively provides a signed distance field \textit{at any scale}. Notably, we define a simple yet effective \textit{inductive bias}, i.e. L2-norm, for regularizing our deep signed distance model. This regularization allows us to obtain expressive deep \gls{sdf}s that can provide precise distance computation over the object's manifold when operating close to it and meaningful level curves when the robot is far from it.

Our deep implicit distance fields give us three significant benefits: first, we can learn and generalize the distance function from data, making it possible to reliably approximate complex manifolds like humans. Second, we can generate smooth and differentiable distance fields that are particularly well-suited for reactive control or trajectory optimization. Finally, deep neural networks allow fast queries of many points in parallel, allowing precise collision avoidance. Our method can be applied to a wide range of robotic applications, from reactive \gls{wbc} and \gls{hri}, to collision-free navigation of aerial robots, as \gls{redsdf} can satisfy any-scale demands depending on the task at hand.  

We demonstrate the capabilities of \gls{redsdf} when employed in reactive motion generation tasks. In particular, we provide a qualitative comparison of the produced distance fields of \gls{redsdf} against representative deep-based baselines, showcasing the advantages of our method. Next, we integrate \gls{redsdf} in reactive motion generation. Namely, we design controllers for the \gls{wbc} of the bimanual mobile manipulator TIAGo++, and for safe \gls{hri} in a shared workspace scenario in a novel simulated task that integrates real-human demonstrations. Finally, thanks to the advances in real-time human skeleton tracking and 3D shape detection, we demonstrate the real-world performance of \gls{redsdf} for computing safety distances when interacting with a human. Our results show the potential of \gls{redsdf} to become a major component of various robotic applications, where precise safety constraints are needed. 

\section{Related Work}
\gls{sdf}s have been studied extensively in the field of computer graphics \cite{curless1996volumetric} to reconstruct meshes of surrounding objects given some range information from lasers or cameras.
The benefits of \gls{sdf}s are very well related to problems of robot motion planning and control, collision-checking, and obstacle avoidance \cite{ratliff2015understanding}. However, while \gls{sdf}s are locally accurate due to truncation effects, they are challenging to construct from partial observations, i.e., when  the whole shape of an object cannot be determined from a single viewpoint. Curless et al. \cite{curless1996volumetric} proposed a volumetric integration method that sacrifices the full coverage of the space for improving the local updates based on partial observations, preserving the representation of positions and orientations on the surface of objects. Truncating the field at small negative and positive values produces the \gls{tsdf}, in which a point outside the truncated region is located in a narrowband that embeds the surface of the object \cite{canelhas2017truncated, oleynikova2016signed}, which allows for better modeling of sensor noise \cite{saulnier2020information}. \gls{tsdf}s are used overwhelmingly in simultaneous localization and mapping applications \cite{izadi2011kinectfusion,vespa2018efficient}. Differently from \gls{tsdf}s that limit the representation capabilities on points close to the surface, the \gls{esdf} provides a method for assessing the free space rather than a fine obstacle area, which is needed for aerial robot mapping and planning \cite{han2019fiesta}. The most popular trajectory optimization methods rely on \gls{esdf} to represent the environment \cite{mukadam2018continuous,schulman2014motion,ratliff2009chomp}, as they are effective in static environments, but suffers high computation time for real-time deployment in dynamic environments.

The deep learning revolution in computer graphics and vision led to the emergence of novel methods for object instance segmentation and reconstruction \cite{han2020occuseg, xu2019explicit} with immediate applications to robot mapping \cite{xu2019mid,grinvald2019volumetric, mccormac2018fusion++}. Deep implicit functions provide a novel way of training and learning to approximate meshes from point clouds and voxelized data from large simulated datasets \cite{peng2020convolutional}.
Deep\gls{sdf} refers to a shape representation method that employs a \gls{sdf} for training a surface reconstruction network by classifying points as belonging to the surface of a mesh or not \cite{park2019deepsdf}. The \gls{ecomann} approach~\cite{sutanto2020learning} extends the application of implicit functions to approximate generic equality constraints for robot motion planning. However, these results only apply to simple scenes, and the distance field estimates are accurate only locally. Other deep learning advances allow for 3D human shape reconstruction from point-clouds \cite{jiang2019skeleton,bhatnagar2020combining} and RGB frames \cite{kocabas2020vibe}, opening up new possibilities for more precise pose estimation of highly articulated meshes, like humans \cite{loper2015smpl}. While implicit representations have been studied in reconstructing the geometry of articulated objects \cite{deng2020nasa, mu2021sdf, gropp2020implicit, atzmon2020sald}, distances for query points that are not on the object surface may be ill-defined for robot manipulation.

In the context of reactive motion control, the use of \gls{sdf}s is challenging due to high computation times that hinder the necessary high control frequency for operating in dynamic environments. To this end, composite \gls{sdf}s were proposed in \cite{finean2021predicted} for propagating SDFs of moving objects by tracking the occupancy box of those objects in the environment. GPU-accelerated voxels combined with whole-body motion planning were explored in \cite{finean2021simultaneous}. \gls{sdf}s are specified across every link of a robot in \cite{xu2018obstacle} for obstacle avoidance during manipulation, while a specific distance function is presented in \cite{quiroz2019whole} for self-collision avoidance in \gls{wbc}.

For \gls{hri}, detecting human poses and understanding the effective \textsl{safe} workspace of the human is essential. Parting from computer graphics, the modeling of highly articulated meshes and the simulation of natural biomechanical properties of human motion \cite{badler1993simulating} became increasingly interesting for robotics. Human-centric knowledge was naturally transferred to human-robot simulated tasks, where the need for specifying the effective workspace of a human partner when interacting with a robot can be considered as a manipulability area around the human skeleton \cite{vahrenkamp2016workspace}. A common way of representing safety constraints in humans for \gls{hri} tasks is through learning the contour around a human pose projected in a low-dimensional space \cite{papadakis2013social, lewandowski2020socially} with primary applications in social human-aware navigation. For closer interactions, many methods represent human links with bounding spheres, from which proximity queries can be made for adapting robot strategies \cite{corrales2011safe,svarny2019safe}.



\section{Learning \gls{redsdf} and Robot Control}

In what follows, we present our unified vision of distance fields through our method \gls{redsdf}. We show how to learn deep-distance fields for complex articulated objects, such as robotic manipulators and humans. Next, we describe the integration of the learned distance fields in reactive motion generation for robot \gls{wbc} and reactive \gls{hri}. For our method, we assume a target-centric coordinate system centered at the center of mass of the object or on any relevant fixed point, e.g. the pelvis position of a human.

\subsection{Regularized Deep Signed Distance Fields}
\begin{figure}
    \centering
    \includegraphics[width=\columnwidth]{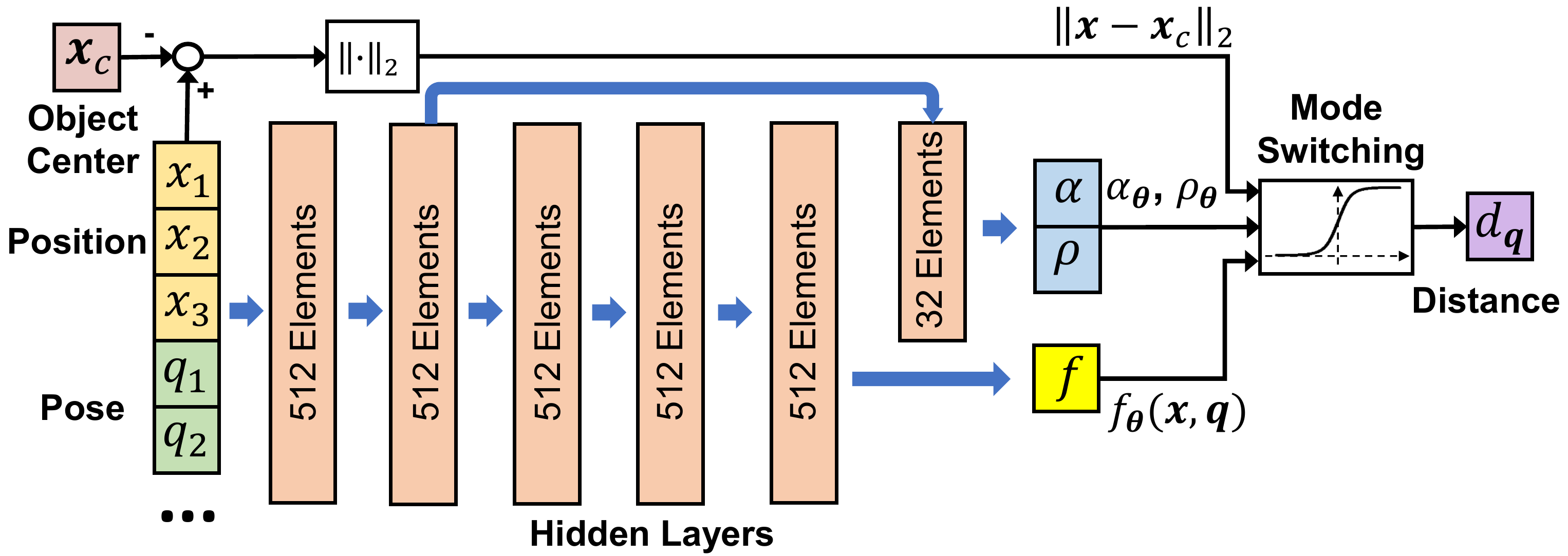}
    \caption{The Network Structure of \gls{redsdf}}
    \label{fig:network}
    \vspace{-0.65cm}
\end{figure}
We exploit two key assumptions to derive a unified vision of \textit{any-scale} deep signed distance fields. First, we assume that the scale of an arbitrary object can be defined using one of its characteristic dimensions, e.g., the radius of the bounding sphere $r$. Thus, for any point $\vx$ where the distance to the center of object $\vx_c$ is $\lVert\vx - \vx_c \rVert_2 \gg r$, the distance is approximated by $d(\vx)\approx \lVert \vx - \vx_c \lVert_2$. Exploiting this assumption makes it possible to build a distance function that unifies most definitions of ``distance'' to a given target. The second key assumption is the dependence of the distance function on the configuration $\vq$ of the considered target. For example, $\vq$ can be the joint positions of a manipulator or the human poses detected by a human tracking system. 

We define \gls{redsdf} w.r.t. a given (articulated) target with configuration $\vq$ at any query point $\vx$ as the distance
\begin{equation}
\small
    d_{\vq}(\vx) = \left(1-\sigma_{\vtheta}(\vx, \vq)\right)f_{\vtheta}(\vx, \vq) + \sigma_{\vtheta}(\vx, \vq)\lVert \vx - \vx_c \rVert_2,
\end{equation}
where $\vtheta$ is the vector of learnable parameters, $\vq$ is the target's state, $f_{\vtheta}(\vx, \vq)$ is a neural network approximator, and the mode switching function $\sigma_{\vtheta}(\vx,\vq)$ is defined as
\begin{equation}
\small
\sigma_{\vtheta}(\vx, \vq)  = \sigmoid\left(\alpha_{\vtheta}\left( \lVert\vx - \vx_c\rVert_2 - \rho_{\vtheta} \right)\right) 
\end{equation}

where $\alpha_{\vtheta}$ and $\rho_{\vtheta}$ are shaping functions, implemented as neural networks. 
Here, the $\rho_{\vtheta}$ function defines a (soft) threshold for switching from a distance w.r.t. the closest point on the target's surface to the distance w.r.t. the center of the target. The $\alpha_{\vtheta}$ function regulates the sharpness of the change between the two regimes.

The network of \gls{redsdf}, depicted in Figure~\ref{fig:network}, is composed of 5 fully-connected hidden layers. All hidden layers are 512-dimensional. In addition, we connect the features from the second layer with a 32-dimensional layer to compute $\alpha_{\vtheta}$ and $\rho_{\vtheta}$. We use ReLU for all layers except for the output layers, which have their specific activation functions. We limit the range of $\rho_{\vtheta} \in (0.5, 1.5)$ by applying a Sigmoid function and a bias on the output unit. $\alpha_{\vtheta}$ is constrained to be positive using a Softplus activation. The distance output $f_{\vtheta}$ is linear.

\paragraph{Data generation and augmentation}
To train \gls{redsdf} w.r.t. a given (articulated) target, we need to generate an appropriate dataset. 
We first obtain a point cloud of the object of interest from different viewpoints and at different configurations $\cmp{\vq}{k}$. Using the Open3D library~\cite{Zhou2018}, we estimate the normal direction $\cmp{\bar{\vn}_{i}}{k}$ for each point $\cmp{\vx_i}{k}$ in the point cloud. The normals are used to augment the original point cloud to different distance levels and to regularize the gradient direction of the \gls{redsdf} as shown in the loss function~\eqref{eq:loss}. However, the estimation of normal directions may be wrong: training can be unstable if the estimated normals don't match the ones of the neighborhood. This issue can be solved by removing such points from the dataset. Using the estimated normals, we augment the data following the procedure of~\cite{sutanto2020learning}, i.e., we augment the data points to $M$ different distance levels by $\cmp{\vx_{ij}}{k}=\cmp{\vx_{i}}{k} + \cmp{\bar{\vn}_{i}}{k}\bar{d}_{j}, j=1, 2, \cdots, M$. For each augmented point, we find its closest point from the original point cloud. If the closest point is not the same as the one from which the augmentation starts $\cmp{\vx_{i}}{k}$, we reject this augmented point. Differently from~\cite{sutanto2020learning}, we do not use a fixed step size, but we heuristically select a set of step sizes, considering the scale of the object. Finally, we assign to every generated point $\cmp{\vx_{ij}}{k}$ a weight $\cmp{\omega_i}{k} = M / \cmp{M_{s, i}}{k}$, where $\cmp{M_{s,i}}{k}$ is the number of the augmented points starting from $\cmp{\vx_i}{k}$ that are not rejected. 
The set of points generated by this method can be unnecessarily large to train a good distance field. To reduce the dataset without losing precision, we uniformly down-sample a subset of the augmented dataset. Finally, we obtain the dataset
\begin{equation}
\small
\mathcal{D}=\lbrace \cmp{\vx_{ij}}{k}, \cmp{\vq}{k}, \bar{d}_{j}, \cmp{\bar{\vn}_{i}}{k}, \cmp{\omega_{i}}{k} \rbrace.
\end{equation}

\paragraph{Human-based data generation}
An important factor, when considering distance to humans, is the high variability of body shapes. This is particularly important when we need to perform close quarter interaction, e.g., for handovers and shared workspace interactions. Instead of using skeleton-based human reconstruction, we leverage the realistic 3D model of the human body, \gls{smpl} \cite{loper2015smpl}. Human 3D meshes are obtained directly from the \gls{smpl} model, given the human poses. The points and normals are extracted from the triangular mesh-elements by computing the center of each triangle and the cross product of the edges. We generate the augmented data as mentioned in previous section. We use the AMASS \cite{AMASS:ICCV:2019} dataset to build up a comprehensive dataset that contains various human poses. In this work, we choose $10,000$ human configurations from the AMASS dataset.

\paragraph{Training}
We train the model by optimizing the following loss function
\begin{align}
\small
\mathcal{L}(\mathcal{D}) = \sum_{\mathcal{D}} &  \omega_{\vq}(\vx)\left(\bar{d}_{\vq}(\vx) - d_{\vq}(\vx)\right)^2 \nonumber\\
& + \left( \lVert D_{\vq}(\vx)\bar{\vn}_{\vq}(\vx)\rVert^2_2 + \lVert N_{\vq}(\vx)\nabla_{\vx} d_{\vq}(\vx)\rVert^2_2 \right) \nonumber\\
& + \gamma\rho_{\vtheta}(\vx,\vq)^2, 
\label{eq:loss}
\end{align}
where $D_{\vq}(\vx) = \mathrm{null}\left(\nabla_{\vx} d_{\vq}(\vx)\right)$ and $N_{\vq}(\vx) = \mathrm{null}\left(\bar{\vn}_{\vq}(\vx)\right)$
are, respectively, the null-space of the gradient of the distance field i.e., the space tangent to the isolines at every point $\vx$, and the null space of the normals i.e., the tangent planes to the object.
The first component of the loss is computing the squared error of the network distance prediction w.r.t. the target value. 
The second component of the loss is similar to the one proposed by~\cite{sutanto2020learning} to align the estimated normals $\bar{\vn}_{\vq}(\vx)$ with the normals of the \gls{redsdf} model.
The last component of the loss is a regularization term, where $\gamma$ is a regularization coefficient, set in our experiments to $0.02$. This regularization is trying to impose the inductive bias of the distance to every point of the state-space.
Reducing the output of the learnable function $\rho_{\vtheta}(\vx,\vq)$ has the effect of switching the distance regime of the network as soon as possible. This regularization is important in particular where the dataset is sparse. The dataset is split into training, validate and test datasets with the ratio $0.8, 0.1, 0.1$.

\subsection{Robot Motion Control with ReDSDF}
\gls{redsdf} can be readily employed within any control and planning framework to provide real-time constraints' inference. Reactive motion generation provides a nice framework to showcase the benefit of our deep-distance field for robot control. Namely, we will describe the integration of \gls{redsdf} in a framework for whole-body motion control based on \gls{apf} \cite{1087247}. Note that \gls{redsdf} can be integrated to any other type of reactive motion generation, such as \gls{rmp} \cite{ratliff2018riemannian}, CEP\cite{Urain2021Composable}. \gls{redsdf} is not restricted to reactive motion generation, but can be employed by any constrained planner, both as a differentiable constraint and as an energy function. For the sake of simplicity, we focus on \gls{apf}-based reactive control, as it directly allows showing the benefits of \gls{redsdf}.

We consider the settings where the \gls{apf}s compute a velocity field for the robot's end-effector. We define a simple PID controller for tracking the end-effector's velocity, but more sophisticated options can be used to generate the task-oriented velocity signal. On top of this, we add an obstacle avoidance velocity signal. Since our method exploits deep neural networks, it is possible to rapidly compute the whole set of collision points in a batch. 
To perform precise obstacle avoidance on a bimanual mobile manipulator robot of Fig.~\ref{fig:sim_exper}.a, we evaluate the \gls{redsdf} over multiple points sampled on the shell of the arms w.r.t. the obstacle. It is noteworthy that the target obstacle can as well be the other arm or any other part of the articulated body of the robot, allowing us to perform self-collision avoidance. For avoiding self-collisions with the other arm of a bimanual robot, we learn a \gls{redsdf} model of the robot with only a single arm. The learned \gls{redsdf} can be used to avoid obstacle for the other arm. We leverage the symmetry of the robot to construct the \gls{redsdf} for the untrained arm. 

For each target obstacle $o$, we compute the obstacle avoidance energy field as
\begin{equation}
\small
    E_{o}(\vx) = \begin{cases}
        0 & d_{\vq}(x) > \kappa \\
        \dfrac{\bar{v}}{2\kappa}\left(d_{\vq}(\vx) -\kappa\right)^2 & 0 \leq d_{\vq}(x) \leq \kappa
    \end{cases},
\end{equation}%
where $\kappa$ is the maximum obstacle avoidance distance, and $\bar{v}$ is the repulsive velocity coefficient.
For $0 \leq d_{\vq}(x) \leq \kappa$, we obtain the obstacle avoidance velocity as follows

\begin{equation}
\small
    \dot{\vx}_o = -\nabla_{\vx}E_o(\vx) = -\dfrac{\bar{v}}{\kappa}\left(d_{\vq}(x) -\kappa\right)\nabla_{\vx}d_{\vq}(\vx).
    \label{eq:repulsive_velocity}
\end{equation}%
While we could also define the energy for negative distance, we choose not to do it as we are considering the rigid obstacle avoidance task, as a negative distance implies a collision with the obstacle. In case of collision, we stop the robot from taking any other action.

\begin{figure*}[t!]
\setlength{\tabcolsep}{0pt}
\renewcommand{\arraystretch}{-1}
\centering
\begin{tabular}{ c c c c c }
    \raisebox{3\normalbaselineskip}[0pt][0pt]{\rotatebox[origin=c]{90}{\textbf{\gls{ecomann}}\cite{sutanto2020learning}}} &
    \includegraphics[width=\imagewidth]{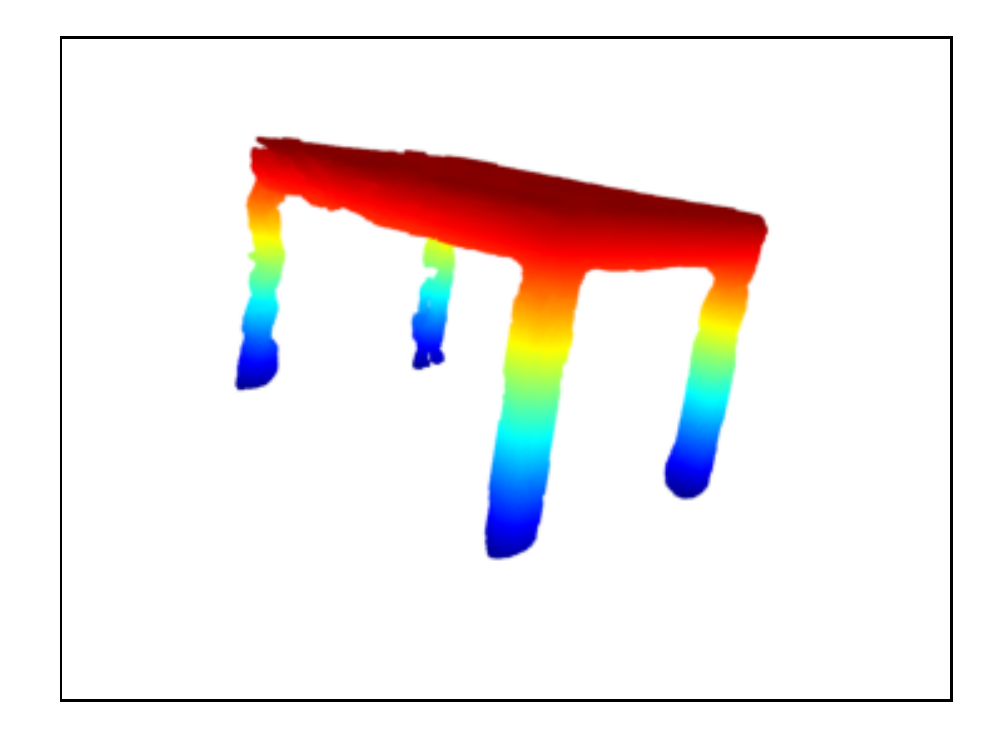} &
    \includegraphics[width=\imagewidth]{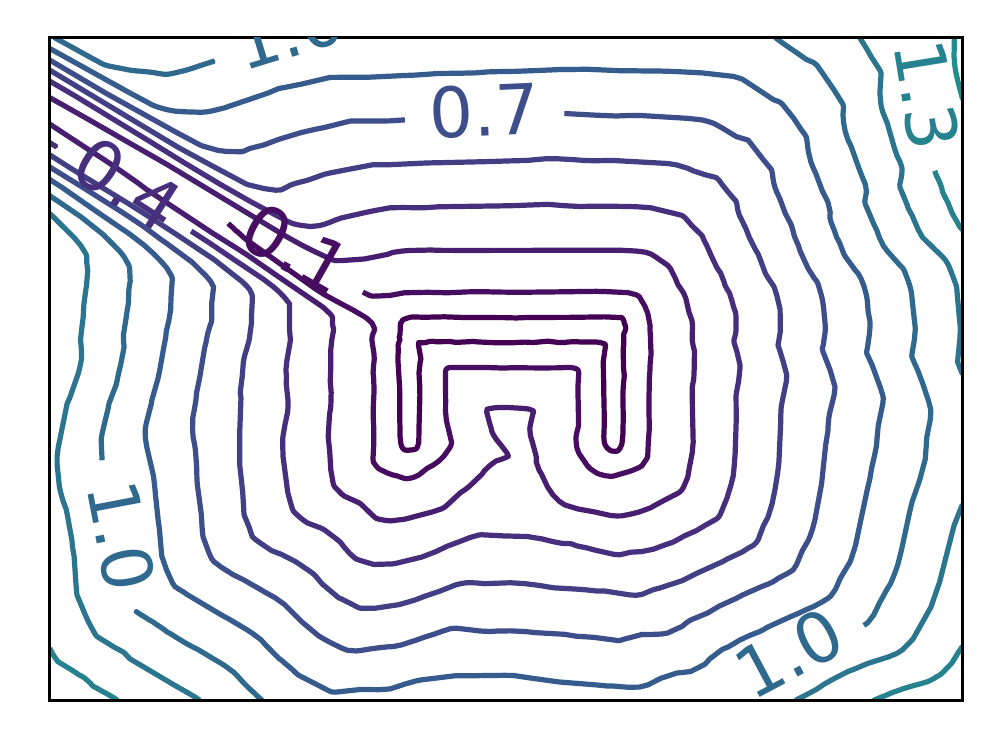} &
    \includegraphics[width=\imagewidth]{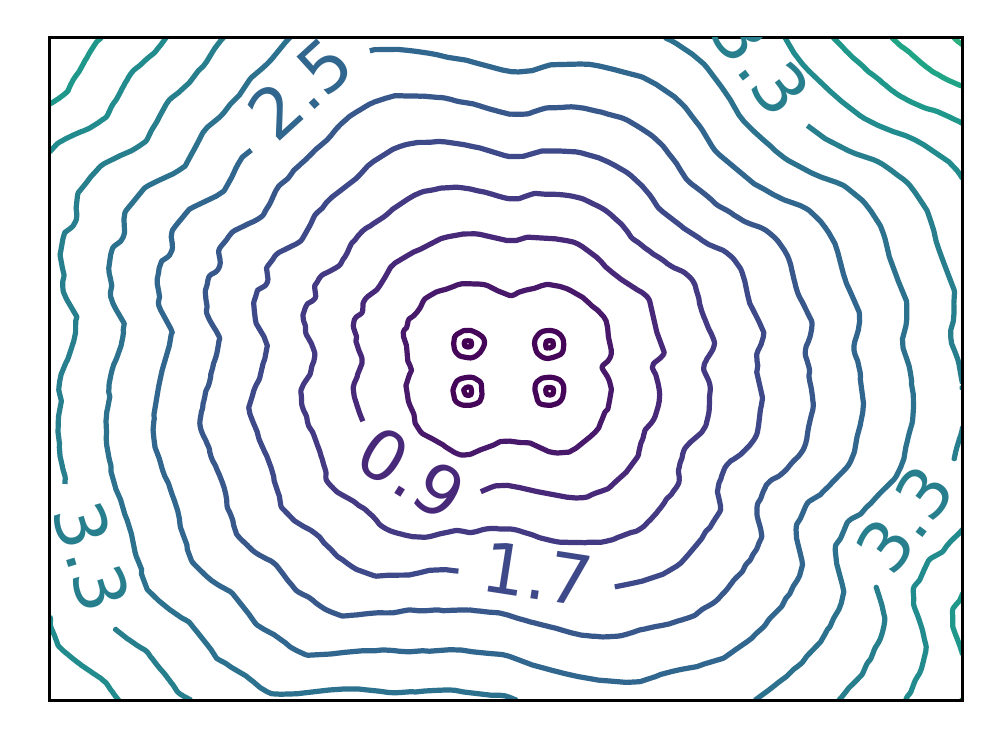} &
    \includegraphics[width=\imagewidth]{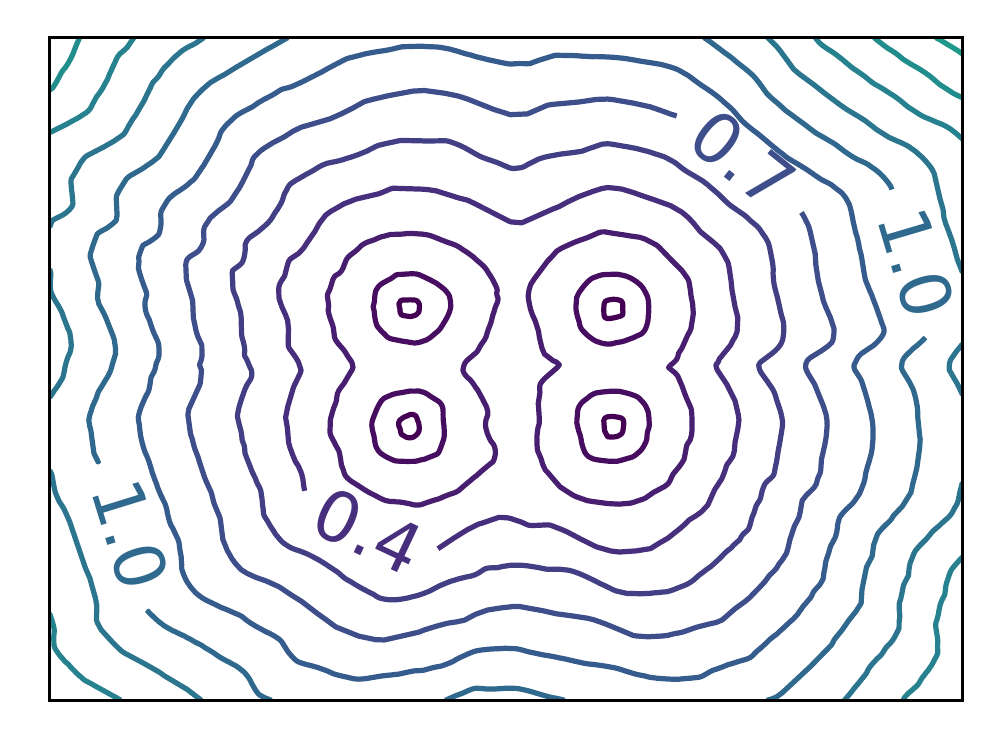}\\
    
    \raisebox{3\normalbaselineskip}[0pt][0pt]{\rotatebox[origin=c]{90}{\textbf{DeepSDF}\cite{park2019deepsdf}}} &
    \includegraphics[width=\imagewidth]{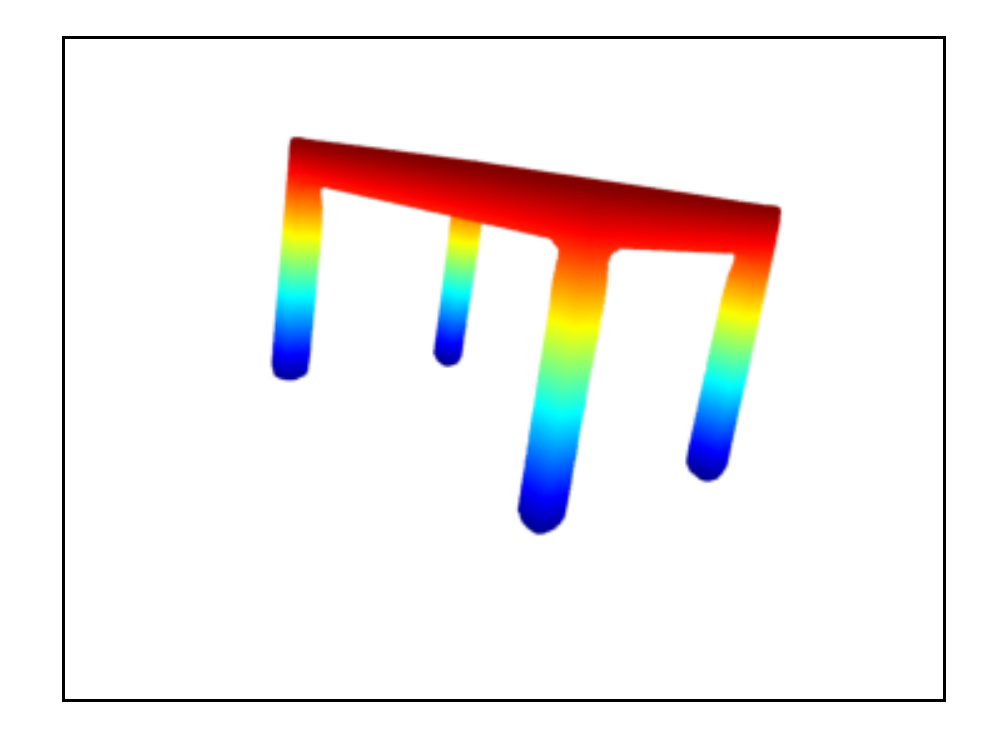} &
    \includegraphics[width=\imagewidth]{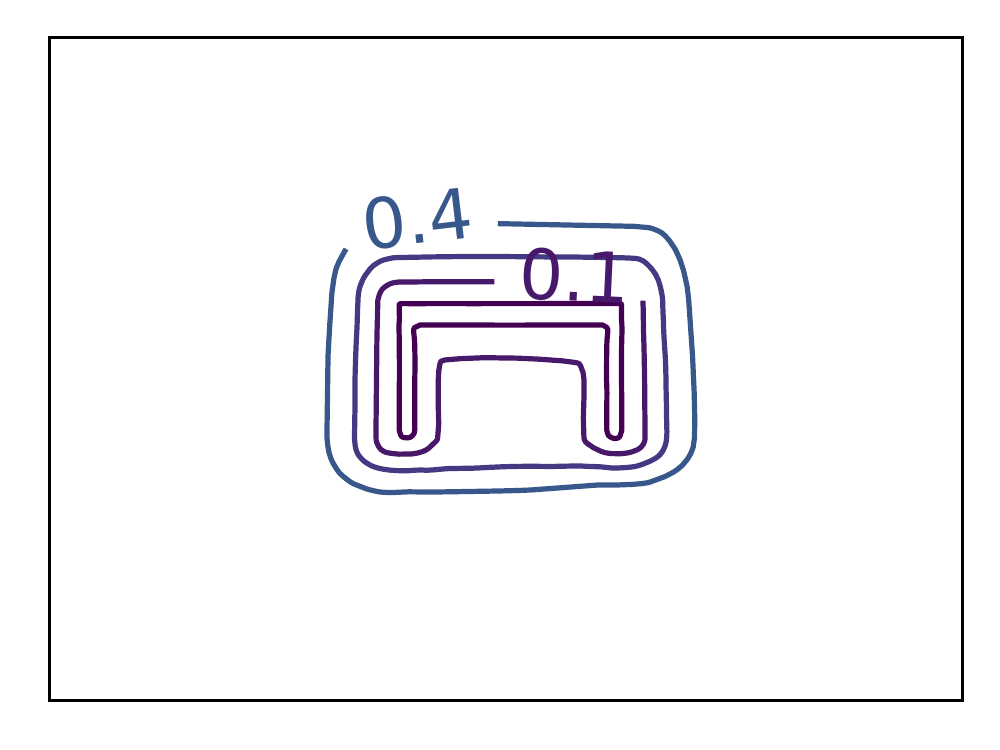} &
    \includegraphics[width=\imagewidth]{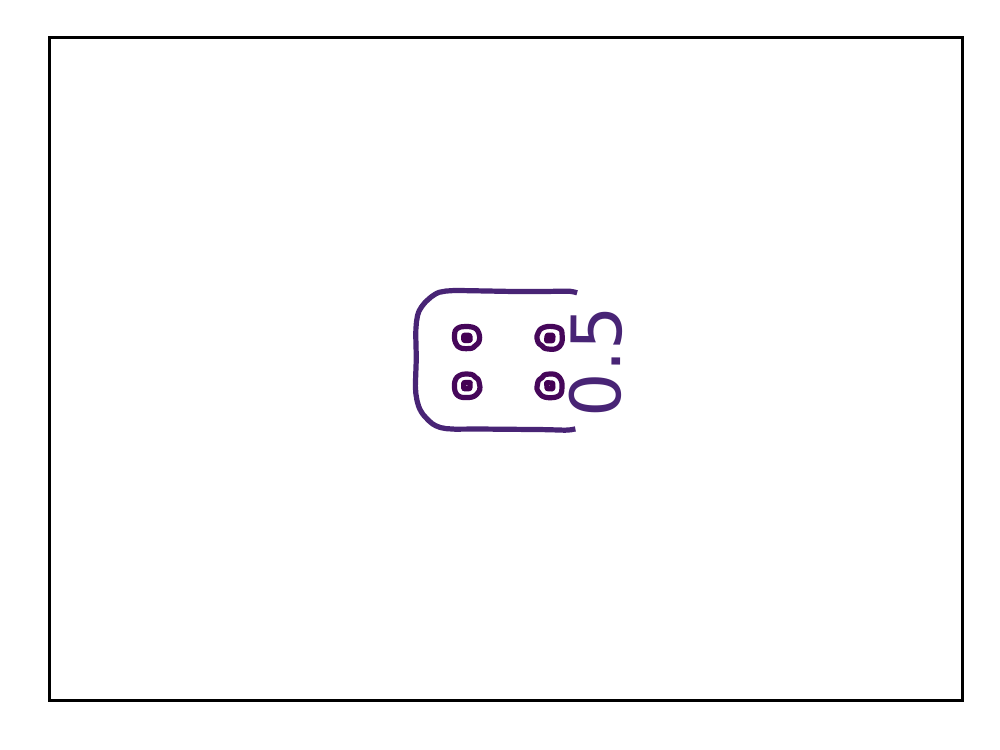} &
    \includegraphics[width=\imagewidth]{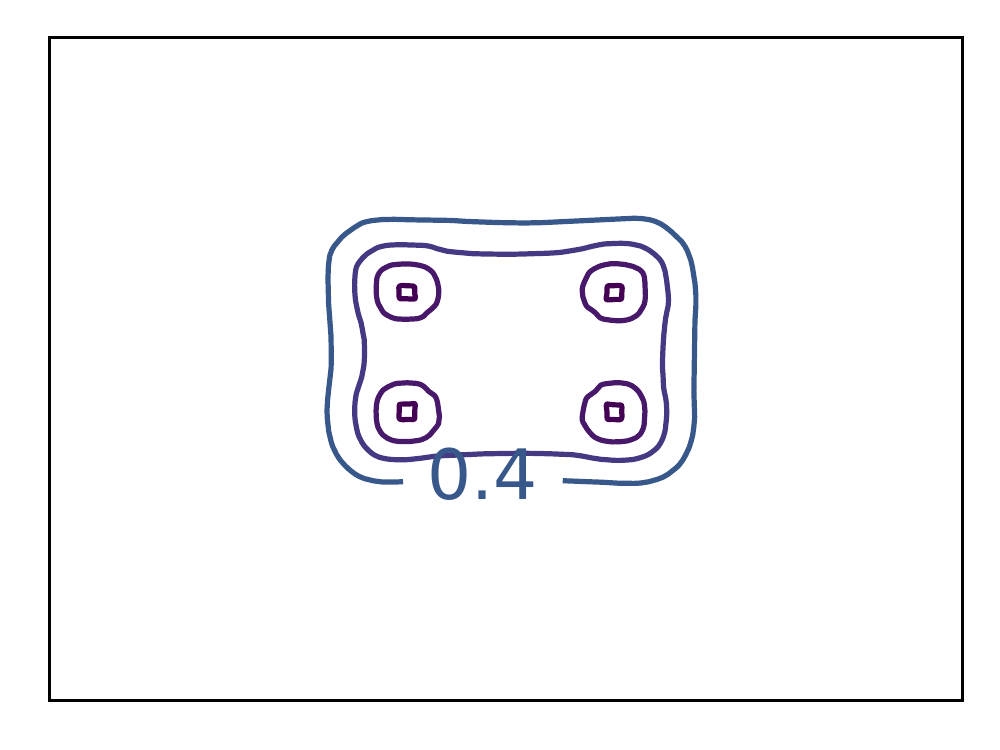} \\
    
    \raisebox{3\normalbaselineskip}[0pt][0pt]{\rotatebox[origin=c]{90}{\textbf{\gls{redsdf}}}} &
    \includegraphics[width=\imagewidth]{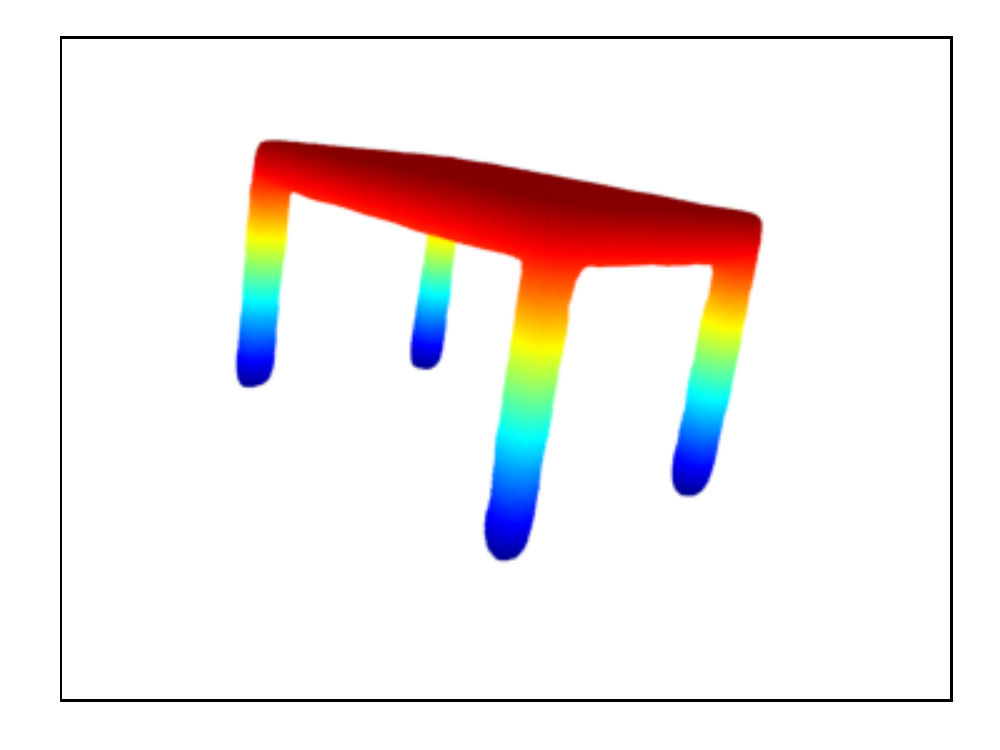} &
    \includegraphics[width=\imagewidth]{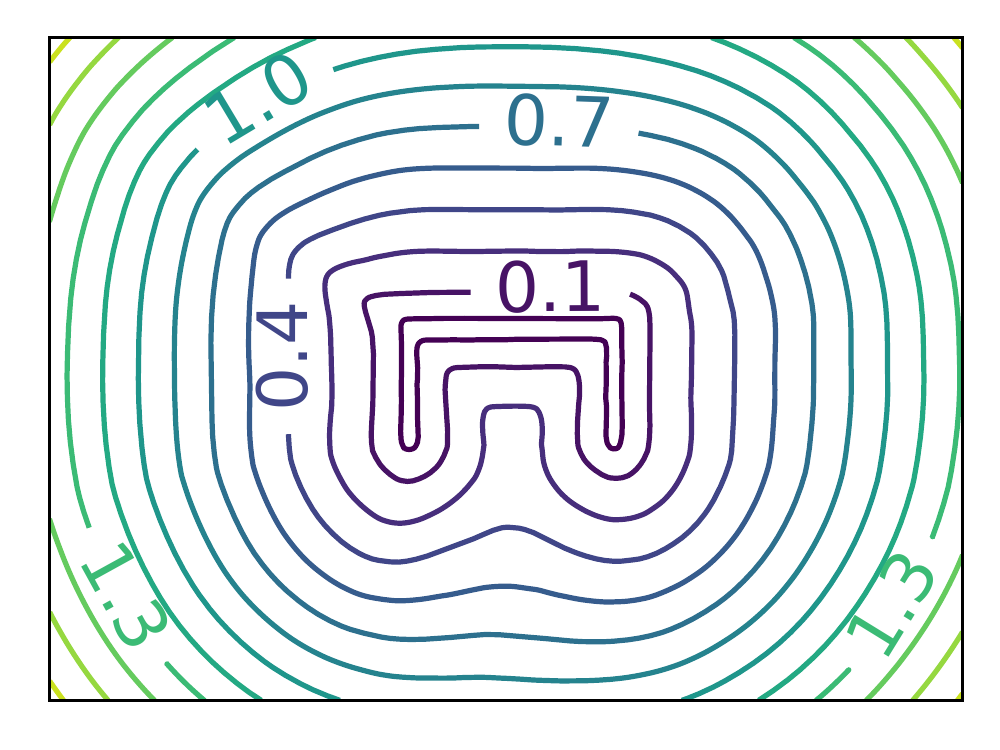} &
    \includegraphics[width=\imagewidth]{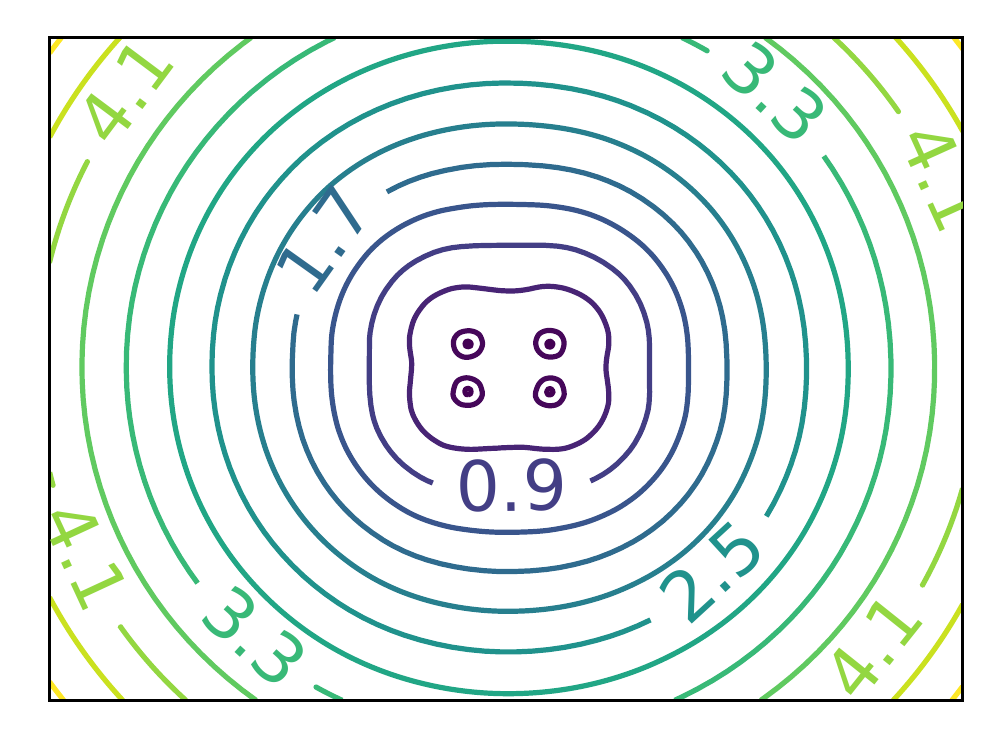} &
    \includegraphics[width=\imagewidth]{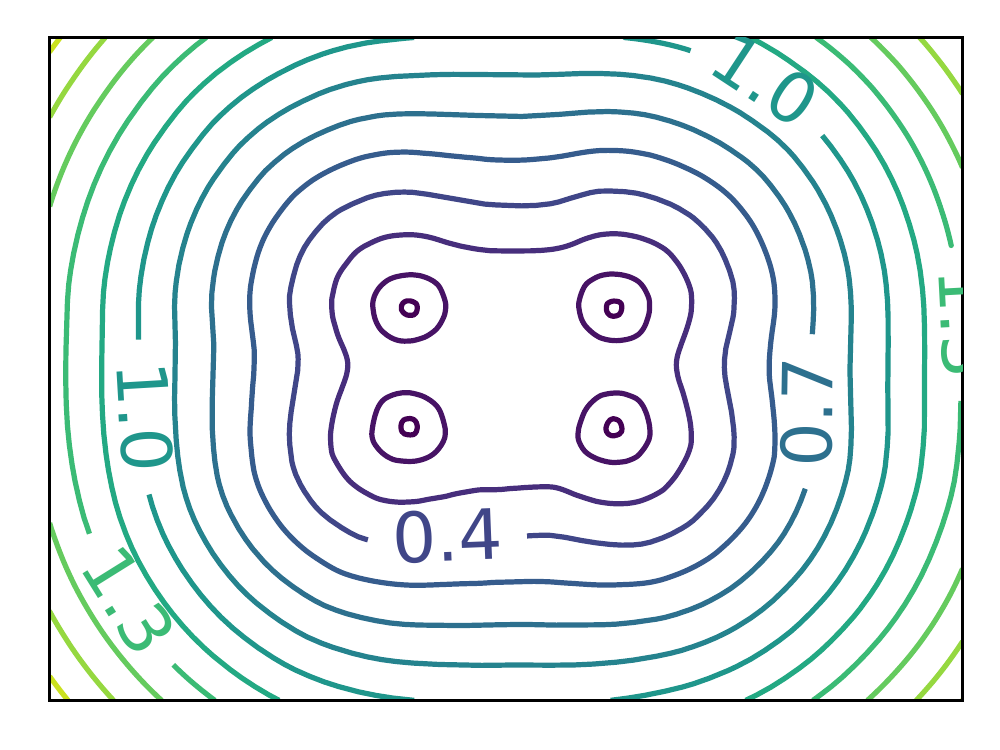} \\
    & \small\textbf{0-level mesh} & \small\textbf{y plane} & \small\textbf{z plane} &  \small\textbf{z plane (zoom)} \\ \\
  \end{tabular}
  \caption{Reconstruction of the table distance function}
  \label{fig:table_manifold}
  \vspace{-0.5cm}
\end{figure*}

The velocity at each \gls{poi} $\vx_i$ is computed by combining the repulsive velocity contribution from the set of obstacles $\mathcal{O}$ in the environment:
\begin{equation}
\small
    \dot{\vx}_i = \dfrac{1}{\lvert\mathcal{O} \rvert}\sum_{o\in \mathcal{O}} \dot{\vx}_{i,o}.
\end{equation}
To transform this velocity from task-space velocity to the robot configuration space velocity $\vq_r$, we can use the Jacobian matrix $J_i(\vq_r)$ computed in the coordinate system defined in $\vx_i$. While the most common solution is to project the velocity into the configuration space using the Jacobian pseudoinverse $J^\dagger(\vq_r)$, in this work we use the Jacobian transpose$\dot{\vq}_{r,i} = J_i^{\intercal}(\vq_r)\dot{\vx_i}.$
This choice is well-known in the literature~\cite{chiaverini1991control,buss2004introduction}, and it is often used to avoid issues with singularities without losing the convergence guarantee.
Finally, we compute the joint velocity by combining the obstacle avoidance velocity with the task velocity $\dot \vx_t$ as
\begin{equation}
\small
\dot{\vq_r} = J_t^{\intercal}(\vq_r)\dot{\vx}_t + \sum_i^N \dot{\vq}_{r,i},
\end{equation}%
where $N$ is the number of interest points and $J_t(\vq_r)$ is the Jacobian computed in the end-effector frame. Instead of using clipping to enforce the joint velocity limits, we rescale all joints' velocities with an appropriate constant, maintaining the same task-space direction while keeping every joint inside its limit. 

\section{Experiments}
To evaluate the effectiveness and applicability of \gls{redsdf} to produce smooth distance fields for robot control, we perform qualitative and quantitative comparisons with various baseline methods in a set of simulated control tasks. First, we evaluate the quality of the produced distance fields of \gls{redsdf} against state-of-the-art methods. Next, we test our method as a component of reactive motion generation for \gls{wbc} and robot control in a shared human-robot workspace. Last, we empirically evaluate the performance of \gls{redsdf} in a realistic \gls{hri} scenario, as seen in Fig. \ref{fig:real_world_interaction}, whose results can be found in the attached video and in our project site \url{https://irosalab.com/2022/02/28/redsdf/}.

\subsection{Quality of the distance field}
\begin{figure*}[t]
\centering
\setlength{\tabcolsep}{0pt}
\renewcommand{\arraystretch}{-1}
\begin{tabular}{ c c c c c }
    \raisebox{3\normalbaselineskip}[0pt][0pt]{\rotatebox[origin=c]{90}{\textbf{\gls{ecomann}}\cite{sutanto2020learning}}} &
    \includegraphics[width=\imagewidth]{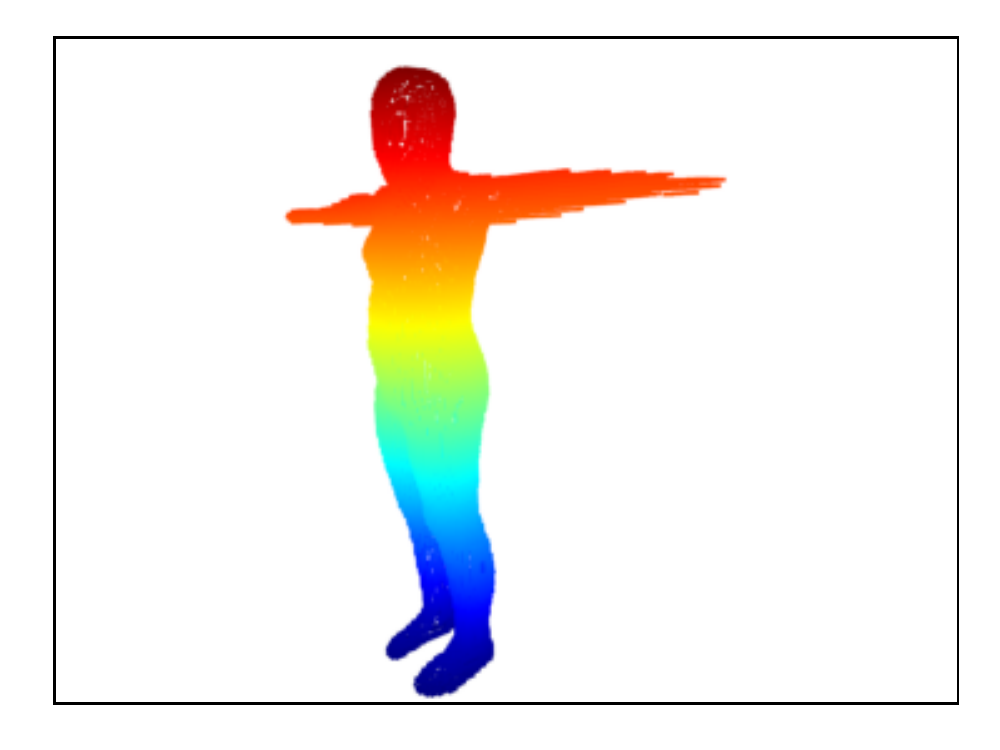} &
    \includegraphics[width=\imagewidth]{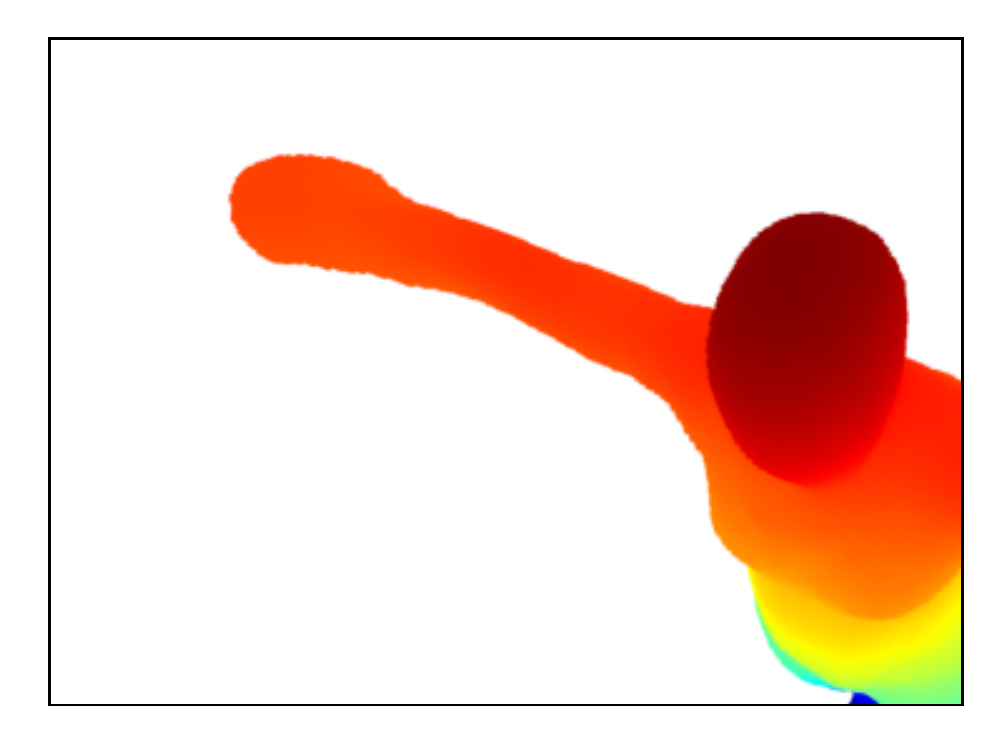} &
    \includegraphics[width=\imagewidth]{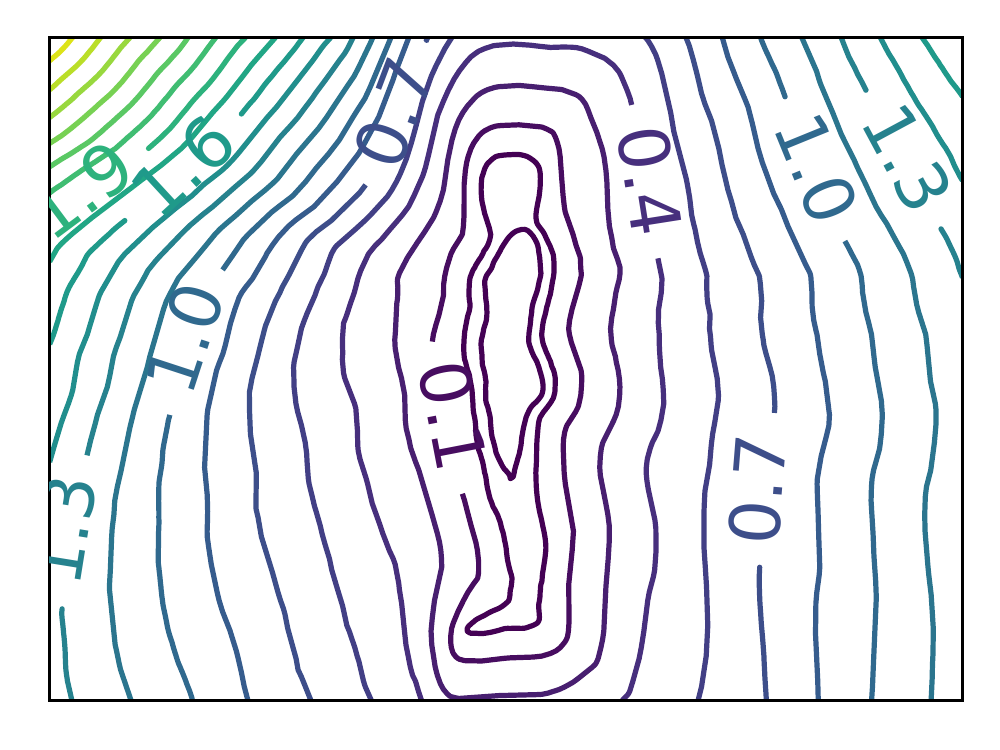} &
    \includegraphics[width=\imagewidth]{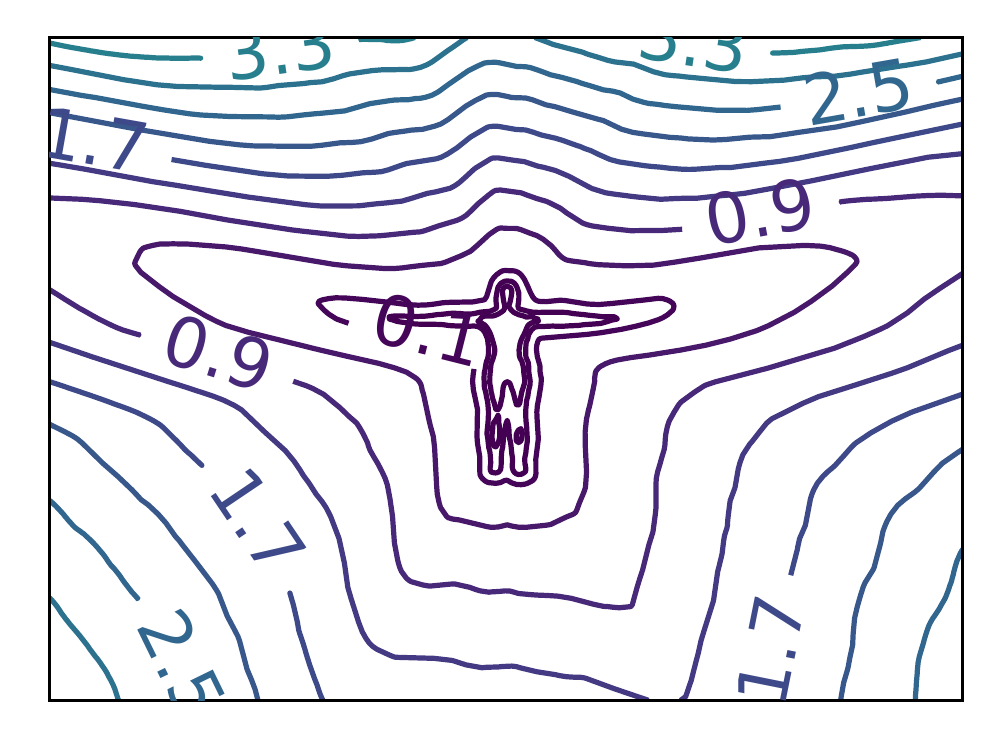}\\
    \raisebox{3\normalbaselineskip}[0pt][0pt]{\rotatebox[origin=c]{90}{\textbf{DeepSDF}\cite{park2019deepsdf}}} &
    \includegraphics[width=\imagewidth]{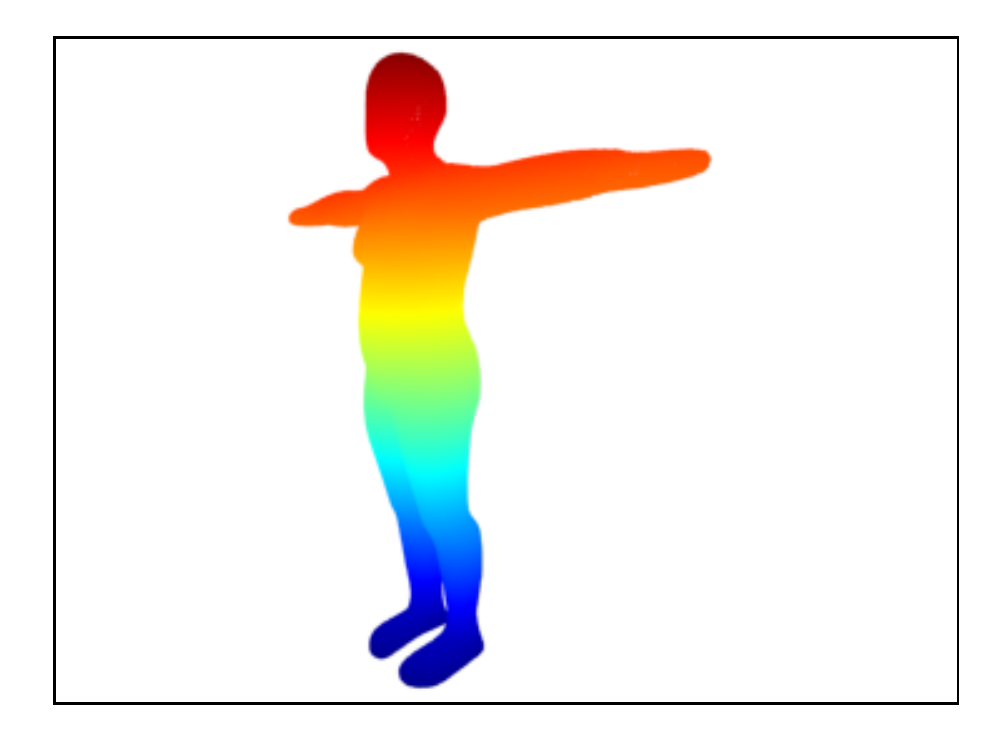} &
    \includegraphics[width=\imagewidth]{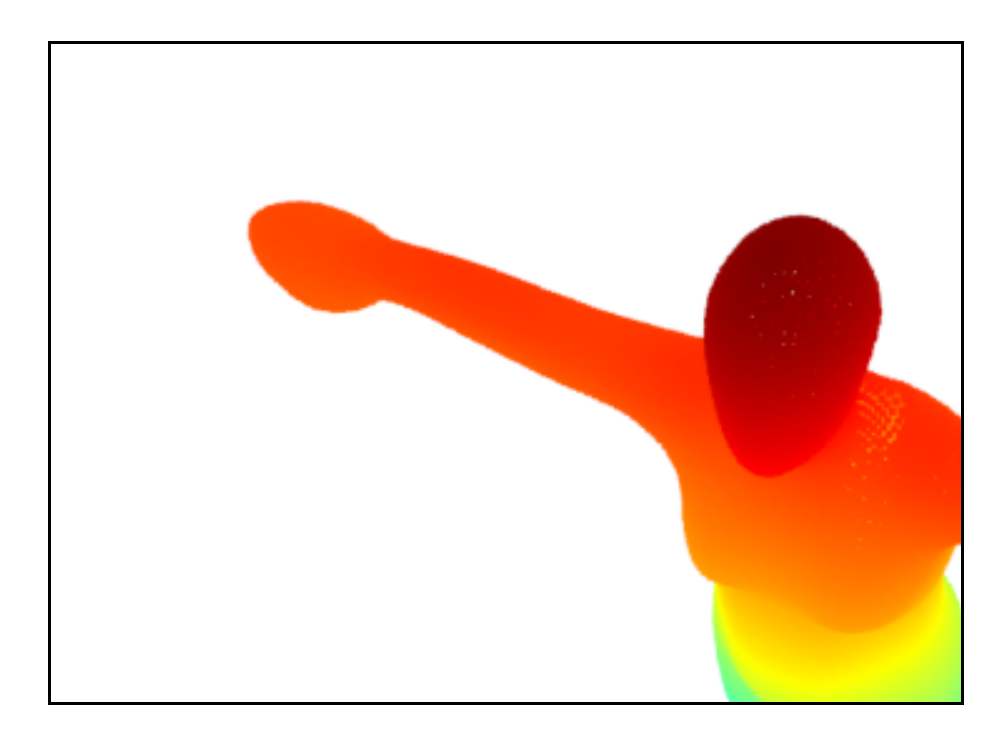} &
    \includegraphics[width=\imagewidth]{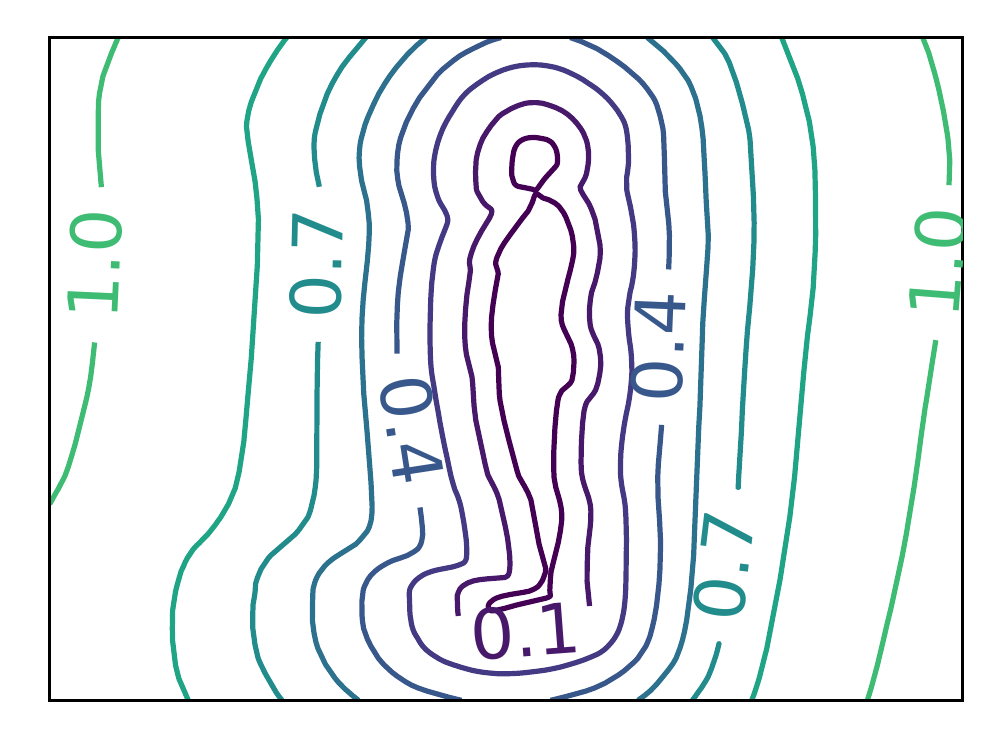} &
    \includegraphics[width=\imagewidth]{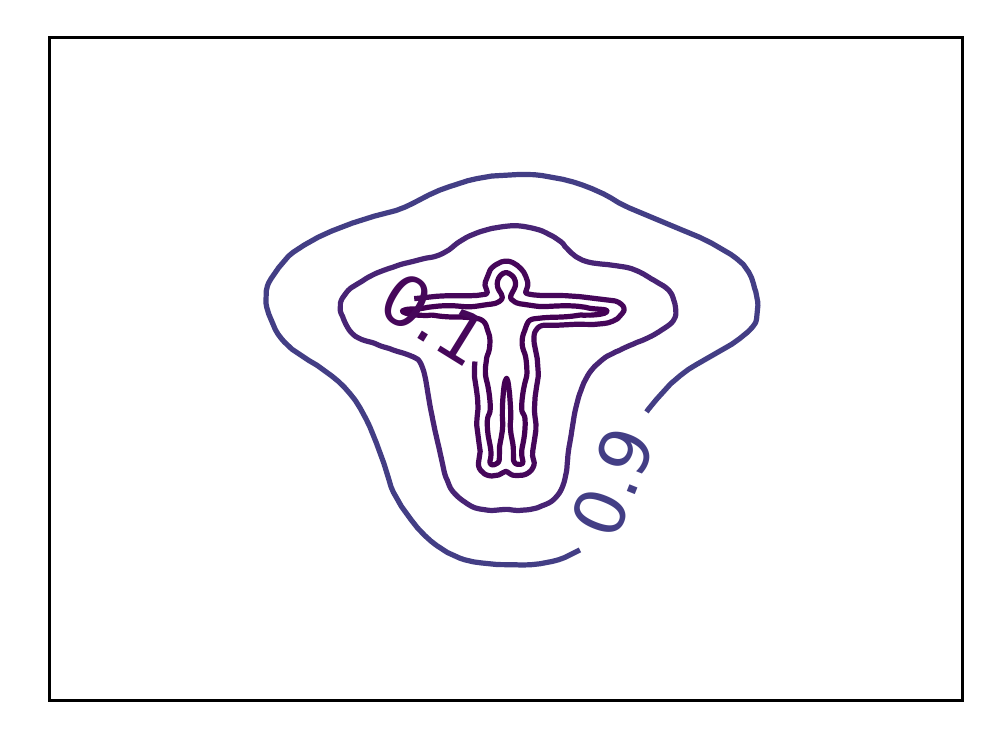}\\
    \raisebox{3\normalbaselineskip}[0pt][0pt]{\rotatebox[origin=c]{90}{\textbf{\gls{redsdf}}}} &
    \includegraphics[width=\imagewidth]{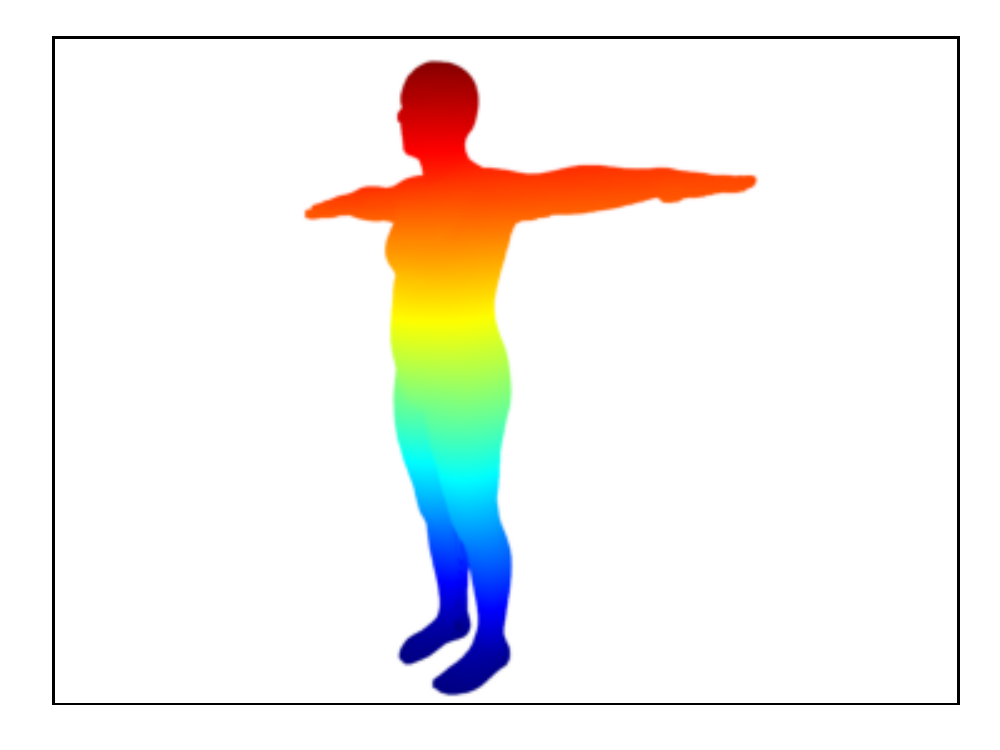} &
    \includegraphics[width=\imagewidth]{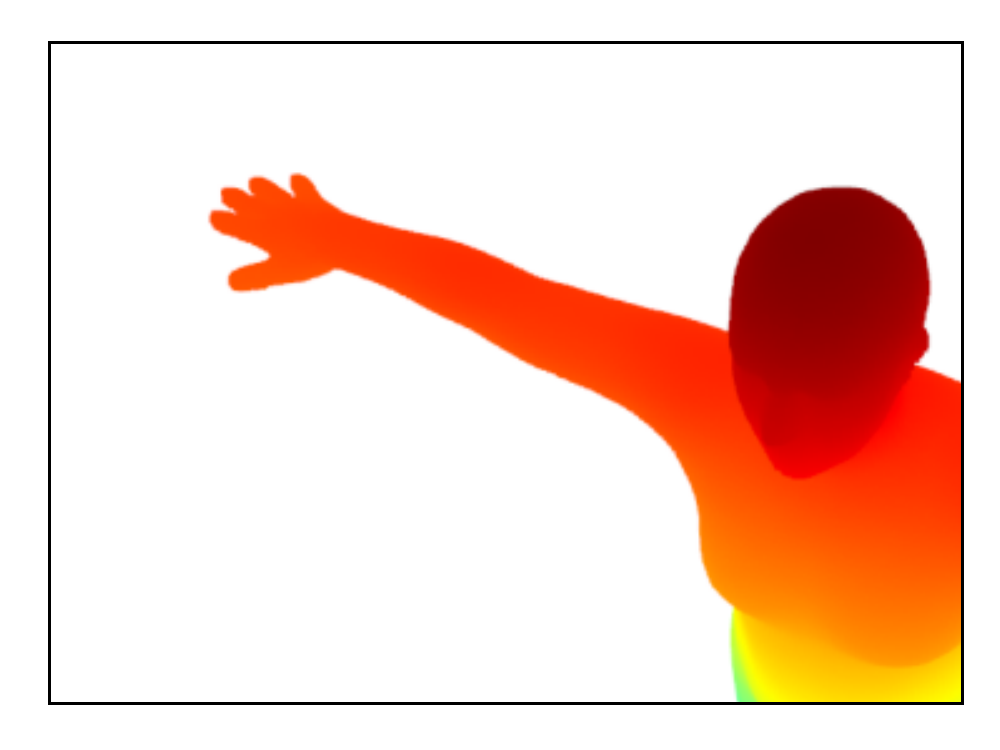} &
    \includegraphics[width=\imagewidth]{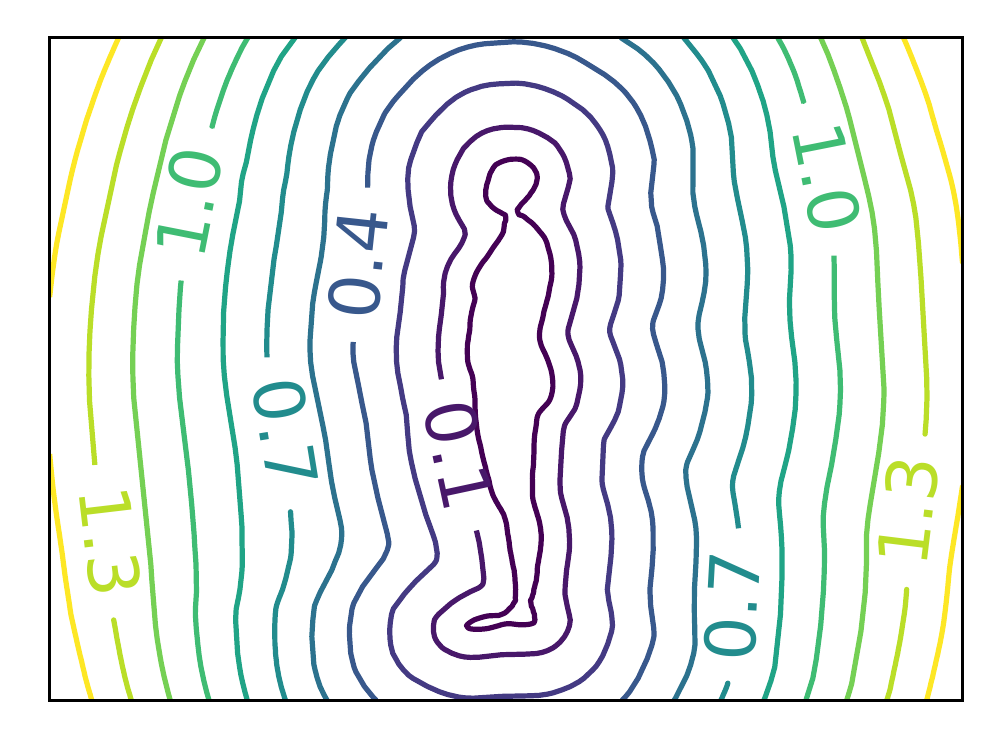} &
    \includegraphics[width=\imagewidth]{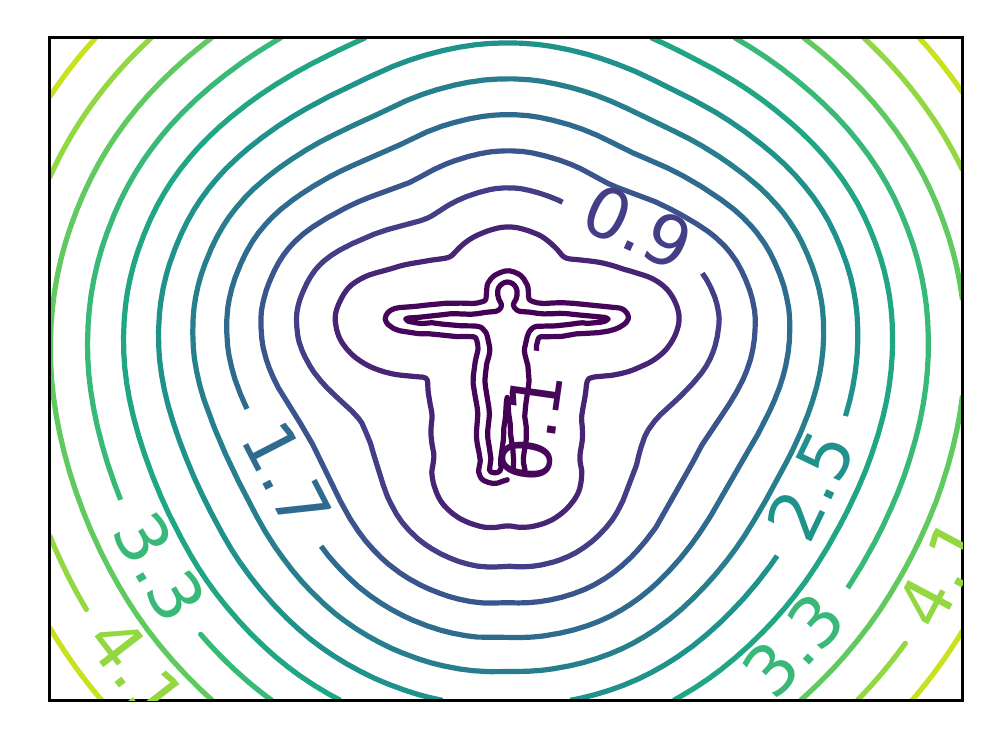} \\
    & \small\textbf{0-level mesh} & \small\textbf{0-level mesh (closeup)} & \small\textbf{x plane (zoom)} & \small\textbf{y plane} \\ \\
  \end{tabular}
  \caption{Reconstruction of the human distance function}
  \label{fig:human_manifold}
  \vspace{-0.5cm}
\end{figure*}

We compare \gls{redsdf} against the state-of-the-art methods \gls{ecomann} \cite{sutanto2020learning} and DeepSDF \cite{park2019deepsdf}. As both baselines were introduced for approximating the manifold for static objects or sampled points cloud, we compare against them with a static object, e.g., a table in Fig.~\ref{fig:table_manifold} and a human mesh with fixed pose Fig.~\ref{fig:human_manifold}. For the baseline approaches, we directly apply the available source code provided by the authors. We use the same databases for all methods, but the training data are generated using the approach-specific data augmentation technique. As the computation of an accurate metric for this task is challenging, we will only provide a qualitative evaluation. However, we argue that the performance difference w.r.t. other state-of-the-art methods is clear, and thus this evaluation is already sufficiently convincing that \gls{redsdf} can provide more rational distance fields for robotics environments.

Fig.~\ref{fig:table_manifold} depicts the results for learning the table manifold. As we can see, the \gls{ecomann} reconstruction contains artifacts in the estimation of the $0$-level curve of the constraint, while our approach reconstructs the mesh without issues. The distance field estimation of \gls{ecomann} is problematic due to wrong normal estimations on the manifold. Instead, DeepSDF reconstructs the manifold well. However, DeepSDF normalizes every mesh file individually and restricts the network's output by a tanh, limiting the distance field's output range. Contrarily, we can observe the superiority of \gls{redsdf} in generating not only a clear geometric manifold of the object, but also smoother and cleaner distance fields at any scale.
Next, we compare the distance field reconstruction from a human mesh file in Fig.~\ref{fig:human_manifold}. All methods reconstruct roughly the human shape. However, \gls{redsdf} provides more precise details like the human fingers and ears. The weighting technique significantly improves the reconstruction of the points in cluttered areas. Furthermore, when looking at the shape of the distance field for points far from the center of mass, we can see that \gls{ecomann} unnaturally deforms the space, e.g., stretching the distance along the hands. DeepSDF, on the other hand, suffers the same scaling problem, while our approach exploits the inductive bias to enforce a well-behaved, even if approximate, distance field. 

For \gls{redsdf}, we generate augmented fine-grained data points in the proximity of the target-surface and more sparse data points as we go away from it. This augmentation allows us to have points at any scale, improving the smoothness of the function and obtaining a precise reconstruction of the $0$-level constraint. The inductive bias we use to regularize the distance field plays a major role in the quality of the reconstruction of the distance function at any scale. Indeed, this is one of the reasons why we do not need to generate augmentation points in the whole state space, and we can make them sparser going away from the manifold without losing the smoothness of the approximated field.

\subsection{Whole-body Control}\label{sec:wbc}
\begin{figure*}[t]
\centering
\setlength{\tabcolsep}{23pt}
\renewcommand{\arraystretch}{-1}
\begin{tabular}{ c c }        
    \centering
    \includegraphics[height=0.22\textwidth]{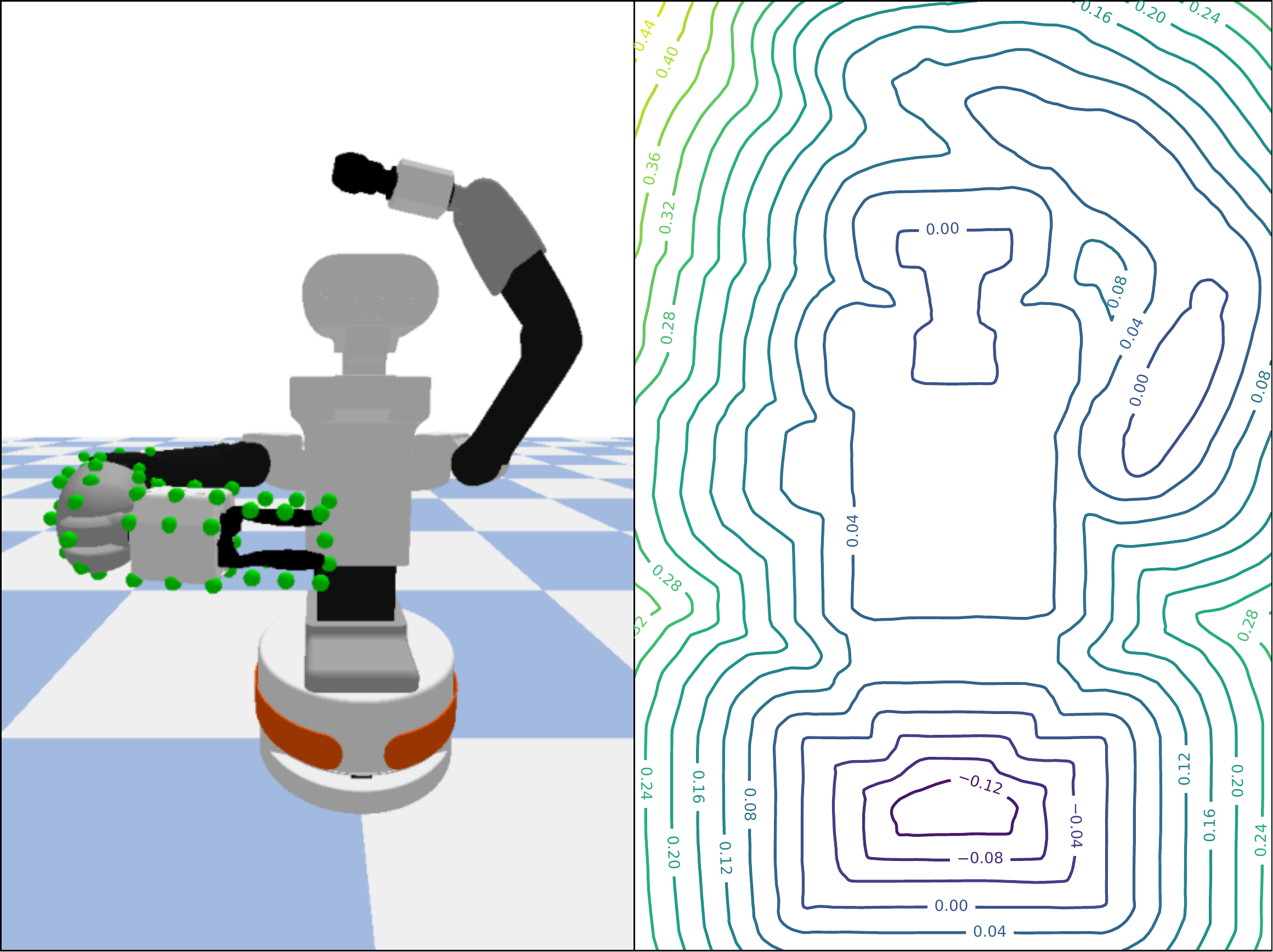} & \frame{\includegraphics[height=0.22\textwidth]{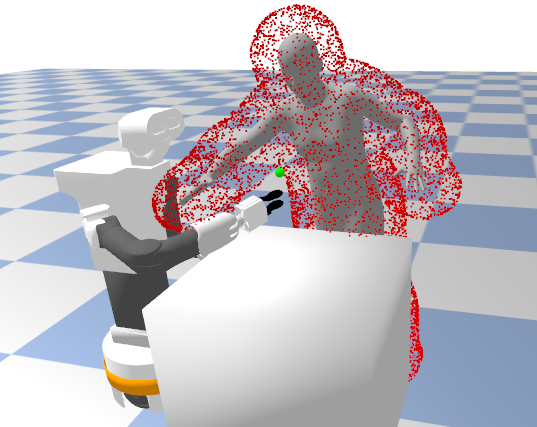}}\\ \rule{0pt}{2.6ex}
    (a) & (b)
\end{tabular}
        \caption{(a) \gls{redsdf} for \gls{wbc}. \textbf{Left:} PoI of the controlled arm that we check to avoid collisions with the rest of the robot. \textbf{Right:} The distance field of the robot, excluding the controlled arm. (b) Human-robot shared-workspace simulated task: human and robot execute sequential pick-and-place-like actions in a tight workspace, where the robot should avoid collisions with the human and the table. The red point cloud represents the $0.1$-level  \gls{redsdf} of the human and is used for collision-avoidance by the robot controller.}
        \label{fig:sim_exper}
        \vspace{-0.2cm}
\end{figure*}
\begin{table*}[t!]
    \centering
    \scriptsize
    \begin{tabular}{ p{1.5cm} c c c c c}
        \multicolumn{6}{c}{\footnotesize\textbf{\gls{wbc} results}} \\\rule{0pt}{2.6ex}
        & \textbf{No avoidance} & \textbf{Sphere-based (task-oriented)} & \textbf{\gls{redsdf} (task-oriented)} & \textbf{Sphere-based (cautious)} & \textbf{\gls{redsdf} (cautious)} \\
        \hline\rule{0pt}{2.6ex}%
        Success rate & 44.8$\%$ & 80.9$\%$ & 82.6$\%$ & 38.3$\%$ & 73.2$\%$\\
        \# collisions  & 548/1000 & 49/1000 & 7/1000 & 0/1000 &  0/1000\\
        Final err. (cm) & 10.42 $\pm$ 0.59 & 1.42 $\pm$ 0.14 & 1.25 $\pm$ 0.11 & 7.21 $\pm$ 0.42 & 1.95 $\pm$ 0.15\\
        Reach time (s) & 6.17 $\pm$ 0.42 & 12.58 $\pm$ 0.58 & 12.74 $\pm$ 0.58 & 23.98 $\pm$ 0.55 & 15.56 $\pm$ 0.65\\
        Smoothness & 10.50 $\pm$ 0.06 & 811.88 $\pm$ 183.40 & 11.62 $\pm$ 0.11 & 3557.51 $\pm$ 284.15 & 12.34 $\pm$ 0.20\\
        C. time (ms) & 0.10 $\pm$ 5.49e-5 & 5.79 $\pm$ 6.28e-3 & 3.27 $\pm$ 5.61e-3 & 5.97 $\pm$ 5.68e-3 & 3.23 $\pm$ 3.18e-3\\
        \hline \\
        \multicolumn{6}{c}{\footnotesize\textbf{Shared Workspace results}} \\\rule{0pt}{2.6ex} 
        & \textbf{No avoidance} & \textbf{Sphere-based (task-oriented)} & \textbf{\gls{redsdf} (task-oriented)} & \textbf{Sphere-based (cautious)} & \textbf{\gls{redsdf} (cautious)} \\
        \hline\rule{0pt}{2.6ex}%
        \# collisions & 935/1000 & 171/1000 & 95/1000 & 86/1000 & 27/1000 \\
        \# targets & 2.172 $\pm$ 0.19 & 6.35 $\pm$ 0.18 & 5.78 $\pm$ 0.16 & 4.44 $\pm$ 0.14 & 4.73 $\pm$ 0.16\\
        Smoothness & 71.95 $\pm$ 0.85 & 2120.00 $\pm$ 198.35 & 135.56 $\pm$ 3.67 & 2286.69 $\pm$ 174.82 & 624.14 $\pm$ 43.50\\
        \hline \\
    \end{tabular}
    \caption{Results of \gls{wbc} and shared workspace experiments in simulation}
    \label{tab:results}
        \vspace{-0.7cm}
\end{table*}
For evaluating the applicability of \gls{redsdf} for robot control, we devise a \gls{wbc} for the bimanual mobile manipulator TIAGo++ robot (Fig. \ref{fig:sim_exper}a). For our simulated experiment, we focus on the control of the two arms and torso (15 degrees of freedom) but assume a stationary base. However, our approach can be trivially extended to generate a base velocity. We generate the augmented dataset of $10,000$ configurations that only contains one arm (left in our experiment). The distance field of the other arm can be computed by mirroring the \gls{poi} through the symmetric plane. We defined $66$ \gls{poi} distributed on the surface of each arm and computed the distance using \gls{redsdf} w.r.t. the other arm and the robot body. Note that the model is learned by excluding each controlled arm from the robot itself; otherwise, every point would be considered in collision. The learned \gls{redsdf} and \gls{poi} of the other arms is illustrated in Fig.~\ref{fig:sim_exper}a.
We use a PID controller to provide the control velocity of the end-effector to reach the target. The repulsive velocity that avoids the collision is computed as described in \eqref{eq:repulsive_velocity}. 
We compare our \gls{wbc} to one without obstacle avoidance and to another one that uses \gls{apf}s but computes the distance using spheres to approximate the robot model, a common approach in the literature. We group the spheres into $3$ subgroups, i.e., body ($18$ spheres), left and right arm ($6$ spheres each). The distances are computed between the spheres from different groups. By adjusting the distance threshold of $\kappa$ and the velocity coefficient $\bar{v}$, we can implement a task-oriented controller that cares more about the end-task, or a cautious controller that places more emphasis on collision avoidance. 

Our experimental results are summarized in the first part of Table~\ref{tab:results}. We conducted $1,000$ experiments with different reaching targets for each arm. To generate the target points, we first sample a random point in space for the first arm, and then we generate a second point for the second arm in the vicinity of the first point. The target points that are too close to the robot are rejected. Each task is simulated for $30$~s with a control frequency of $60$~Hz. The episode terminates once the simulator detects a collision based on the meshes. We regard the task as successful when the end-distance of the end-effector to the target is less than $3$~cm. We report the success rate percentage, the absolute number of collisions, and the mean and confidence interval of the final position error, the reach time, smoothness of the generated trajectory $\tau$ as:
$ s(\tau) = \sum_{\vq_r(i)\in \tau} \lVert\ddot{\vq}_r(i)\rVert^2$, and of the time spent for computing collisions (C. time). 

Due to our design choices, both the sphere-based distance computation and \gls{redsdf} are comparable in terms of success rate when the controller is task-oriented, but \gls{redsdf} only has seven collisions. For the cautious controller, \gls{redsdf} provides zero collisions and satisfactory performance, while the heuristic sphere-based method requires strong repulsive forces to eliminate collisions, hurting the task performance. For the smoothness metric, we consider both changes in the direction of the velocity and changes in magnitude, and for a fair comparison, we only compute the smoothness for the collision-free episodes. \gls{redsdf} generates a continuous distance field, which results in a much smoother controller. For the heuristic, the distance and normal directions are chosen based on the closest sphere of each \gls{poi}. The normals may change drastically when the closest sphere varies, which results in jittery movements. The distance computation time is evaluated on an AMD Ryzen 7 3700X 8-Core Processor with a GeForce RTX 2080 SUPER. We compute the distance of the 132 \gls{poi} (66 \gls{poi} per arm) with \gls{redsdf} in batch on the GPU using PyTorch. For the sphere-based method, 288 distance queries are requested at each time in batch on the GPU to speed up its performance. The computation time for \gls{redsdf} is $50\%$ less than the heuristic. We want to point out that the computation load of \gls{redsdf} grows linearly w.r.t. the number of \gls{poi}, while for the sphere-based method it grows quadratically. This advantage allows us to achieve more query points and finer computations of the distance field.

\subsection{Shared Human-Robot workspace}

To test the applicability of \gls{redsdf} for reactive motion generation for challenging \gls{hri} tasks, we built a simulation task where a human and a robot are performing sequential pick-and-place-like actions in a shared workspace. To simulate realistic movements, we record a set of trajectories from an actual human performing the task, and we infer the \gls{smpl} trajectories with the help of VIBE\cite{kocabas2020vibe}. We replay these trajectories in our PyBullet simulator, constituting a novel way of building simulation tasks for \gls{hri} with the robot dynamically interacting with the human (Fig. \ref{fig:sim_exper}b). Instead of reaching the fixed targets as in \gls{wbc} task, we generate 9 targets for each task to mimic the pick-and-place scenario, a new target will be activated once the old target has been reached. We assume that the human is not responsive to the robot in the current setting. While this assumption is unrealistic, we want to test the reactiveness of the robot controller while interacting in a tight workspace with a human. If the human actively avoids the robot, it would be impossible to test if our controller successfully avoids collisions. Indeed, we can see the proposed scenario as an approximation of a human not paying attention to the robot while performing a shared-workspace task.

Table~\ref{tab:results} contains the results of this experiment. We compared the collision avoidance behavior based on \gls{redsdf} with two simple baselines. The first, trivial baseline is to plan without considering the human at all. The second baseline is a prevalent approach in the literature that considers only the human hand's position and reacts to hand movements. We developed similar task-oriented and cautious controller as in Sec. \ref{sec:wbc}. For all methods, we also include a potential field to prevent collision with the table. Inspecting the results, we can observe that our controllers with \gls{redsdf} can provide smoother distance fields that result in smoother robot movements, with significantly fewer collisions compared to the baseline methods. Even though in the target-oriented case the heuristic achieves more targets, it suffers from a higher number of collisions. Overall, our results validate our assumption that \gls{redsdf} is effective and smooth when used in real-time as a constrained function for safe and reactive \gls{hri}. 
As proof of concept, Fig. \ref{fig:real_world_interaction} and the accompanying video provides a real-world experiment in a realistic \gls{hri} handover scenario, where we can inspect the any-scale smooth performance of \gls{redsdf}.

\section{Conclusions}
Safety is a critical property for autonomous robots, both for planning actions that respect their hardware and the surrounding world and when robots interact with humans in shared workspaces. 
In this work, we presented the \glsfirst{redsdf} framework, which generalizes the concept of \gls{sdf} to arbitrary articulated objects. We regularize the training of the distance function with an inductive bias to ensure smooth distance fields that can be used at any scale. Using the proposed technique, we obtain high-quality, smooth distance fields for any point in the space. Our method can be easily deployed to generate reactive motions both in the context of a \glsfirst{wbc} and in a human-robot shared workspace scenario, using real human data. The reactive motion can reliably avoid obstacles while reducing the interference considerably with the main task, allowing us to outperform baselines methods by a significant margin in many metrics.
The \gls{redsdf} structure can be easily deployed in real robot applications, as it can provide accurate distance fields in different resolutions, from rough distance estimates for obstacle avoidance during navigation to fine-grained resolutions in close interactions with humans. Though we demonstrated the applicability of \gls{redsdf} to reactive motion generation, thanks to its smooth differentiable nature, it can be easily integrated into different pipelines, such as constrained task and motion planning or constrained reinforcement learning.




\bibliographystyle{IEEEtran}
\bibliography{bibliog}

\end{document}

%% file: math.tex
\DeclareMathOperator{\sigmoid}{sigmoid}

\newcommand{\vn}{\bm{n}}

\newcommand{\vq}{\bm{q}}

\newcommand{\vx}{\bm{x}}

\newcommand{\vtheta}{\bm{\theta}}

\newcommand{\cmp}[2]{#1^{[#2]}}

%% file: acronyms.tex
\newacronym{rl}{RL}{Reinforcement Learning}
\newacronym{sdf}{SDF}{Signed Distance Function}
\newacronym{tsdf}{TSDF}{Truncated Signed Distance Function}
\newacronym{esdf}{ESDF}{Euclidean Signed Distance Function}
\newacronym{redsdf}{ReDSDF}{Regularized Deep Signed Distance Fields}
\newacronym{apf}{APF}{Artificial Potential Fields}
\newacronym{rmp}{RMP}{Riemannian Motion Policies}
\newacronym{hri}{HRI}{Human-Robot Interaction}
\newacronym{ecomann}{ECoMaNN}{Equality Constraint Manifold Neural Network}
\newacronym{smpl}{SMPL}{A Skinned Multi-Person Linear Model}
\newacronym{poi}{PoI}{Points of Interest}
\newacronym{wbc}{WBC}{whole-body control}